\documentclass[journal]{IEEEtai}
\usepackage[lined,ruled,commentsnumbered]{algorithm2e}
\usepackage{graphicx,url,subfigure,multirow,booktabs,amsmath,amssymb,threeparttable,bm}

\begin{document}

\title{Semi-Supervised Transfer Boosting (SS-TrBoosting)}

\author{Lingfei~Deng, Changming~Zhao, Zhenbang~Du, Kun~Xia, and Dongrui~Wu
\thanks{L.~Deng, C.~Zhao, Z.~Du, K.~Xia and D.~Wu are with the Ministry of Education Key Laboratory of Image Processing and Intelligent Control, School of Artificial Intelligence and Automation, Huazhong University of Science and Technology, Wuhan 430074, China. They are also with Shenzhen Huazhong University of Science and Technology Research Institute, Shenzhen, 518063, China.}
\thanks{D.~Wu is the corresponding author (e-mail: drwu09@gmail.com).}}

\maketitle

\begin{abstract}
Semi-supervised domain adaptation (SSDA) aims at training a high-performance model for a target domain using few labeled target data, many unlabeled target data, and plenty of auxiliary data from a source domain. Previous works in SSDA mainly focused on learning transferable representations across domains. However, it is difficult to find a feature space where the source and target domains share the same conditional probability distribution. Additionally, there is no flexible and effective strategy extending existing unsupervised domain adaptation (UDA) approaches to SSDA settings. In order to solve the above two challenges, we propose a novel fine-tuning framework, semi-supervised transfer boosting (SS-TrBoosting). Given a well-trained deep learning-based UDA or SSDA model, we use it as the initial model, generate additional base learners by boosting, and then use all of them as an ensemble. More specifically, half of the base learners are generated by supervised domain adaptation, and half by semi-supervised learning. Furthermore, for more efficient data transmission and better data privacy protection, we propose a source data generation approach to extend SS-TrBoosting to semi-supervised source-free domain adaptation (SS-SFDA). Extensive experiments showed that SS-TrBoosting can be applied to a variety of existing UDA, SSDA and SFDA approaches to further improve their performance.
\end{abstract}


\IEEEpeerreviewmaketitle

\section{Introduction}\label{sec:introduction}

A well trained machine learning model may not work as well on a new dataset or in a new environment, due to the large distribution discrepancy between the training data (source domain) and the test data (target domain). Domain adaptation (DA) \cite{wang2018deep} is a feasible solution to this problem. It aims at reducing the distribution differences between the source and target domains for better performance in the target domain~\cite{zhang2019tdl}. DA is especially important when the target data labels are completely missing or insufficient \cite{wilson2020survey}. Most existing DA research focused on unsupervised domain adaptation (UDA), where only source data are labeled. Recent studies \cite{li2018semi,saito2019semi,kim2020attract} showed that a few labeled target data in semi-supervised domain adaptation (SSDA) can significantly improve the DA generalization performance.

\begin{figure*} [htbp]    \centering
    \includegraphics[width=\textwidth,clip]{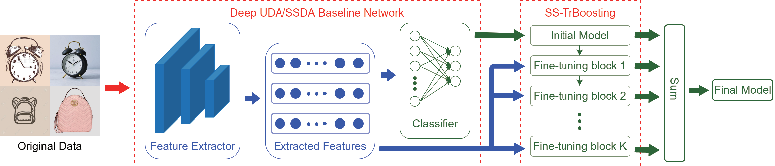}
    \caption{Illustration of the proposed SS-TrBoosting, which fine-tunes the initial deep learning classifier by adding a series of boosting blocks. The same features from the feature extractor are fed into each fine-tuning block as inputs.}     \label{fig:intuition}
\end{figure*}

This paper focuses on two challenges in SSDA:
\begin{enumerate}
    \item Reducing the domain alignment bias, by making use of the labeled target data. Popular SSDA approaches mainly focus on learning transferable feature representations across domains. Unfortunately, the distribution discrepancy between the source and target domains usually cannot be eliminated completely. For example, Minimax entropy (MME) \cite{saito2019semi} and enhanced categorical alignment and consistency learning (ECACL) \cite{li2021semi} reduce the bias by prototype alignment. However, prototypes are representative only when the sample distribution in each class is Gaussian, which may not always hold in practice. Bidirectional adversarial training (BiAT) \cite{jiang2020bidirectional} and attract, perturb, and explore (APE) \cite{kim2020attract} add perturbations to the labeled target data and use adversarial training for DA. However, insufficient labeled target data may lead to biased alignment.

    \item Enhancing the model flexibility, by making use of existing deep-learning-based UDA approaches. Compared with SSDA, UDA has been more extensively studied; however, there are few universal and effective strategies extending existing state-of-the-art UDA approaches to SSDA. Moreover, MME \cite{saito2019semi} and APE \cite{kim2020attract} showed that directly combining labeled data in both domains to train a UDA model may not be effective.
\end{enumerate}

Deep-learning-based DA approaches usually consist of a feature extractor and a classifier. They mainly focus on training a transferable feature extractor. To cope with the above two challenges, we switch the attention from extracting transferable features to generating a transferable classifier. The scenario considered in this paper is: \emph{Given a pre-trained UDA or SSDA model, how to modify its classifier to further improve the generalization performance?}

Boosting \cite{boosting} is a widely used ensemble learning strategy to enhance the generalization performance of a base model. TrAdaBoost \cite{dai2007boosting} combines boosting and DA to reduce the prediction bias caused by distribution discrepancy. However, TrAdaBoost is not compatible with deep-learning-based DA approaches, which limits its applications. We propose a novel approach to combine deep-learning-based DA and boosting to enhance the performance of the former. We take LogitBoost \cite{friedman2000additive} as the ensemble strategy due to its good compatibility with many regression models, e.g. ridge model \cite{ridge1970} and neural networks.

Moreover, due to privacy concerns, labeled source data may not be available in some real-world applications \cite{li2020model}. Therefore, we consider a more challenging scenario: semi-supervised source-free domain adaptation (SS-SFDA), where a trained source model (instead of the source data), few labeled target data and plenty of unlabeled target data are available. There are two common strategies for SFDA: 1) Generate a synthetic domain similar to the original source domain~\cite{li2020model,qiu2021source}; 2) Fine-tune the provided source model by using the target data \cite{liang2020shot,yang2021nrc}.  The first is used in SS-SFDA.

Our main contributions are:
\begin{enumerate}
    \item We propose semi-supervised transfer boosting (SS-TrBoosting), which embeds LogitBoost \cite{friedman2000additive} in deep DA networks for fine-tuning. As shown in Figure~\ref{fig:intuition}, SS-TrBoosting first uses the DA model's feature extractor to pre-align the source and target domains, and takes the DA model's classifier as the initial model. Then, it uses the feature extractor's outputs as inputs to generate a series of fine-tuning blocks, each including two base learners to handle the domain alignment bias and to exploit the unlabeled target data, respectively. Finally, it combines the initial model and all fine-tuning blocks as the ensemble model.

    \item SS-TrBoosting is compatible with a variety of SSDA approaches, and can extend deep-learning-based UDA approaches to SSDA. Experimental results on three benchmark datasets demonstrated that SS-TrBoosting can further improve the generalization performance of state-of-the-art UDA and SSDA approaches.

    \item We propose a novel source data synthesis approach, which generates a virtual source domain, to extend SS-TrBoosting to SS-SFDA to protect the privacy of the source domains.
\end{enumerate}

\section{Related Work}

\subsection{Semi-supervised learning (SSL)}

 SSL uses a large amount of unlabeled data and few labeled data to train a model with good generalization \cite{chapelle2009semi}. The key is how to use the information of unlabeled data. Pseudo labeling uses the predictions of the unlabeled data as the true labels to update the classifiers \cite{hu2021simple}. Entropy minimization minimizes the entropy of model outputs of the unlabeled data \cite{grandvalet2005semi}. Consistency regularization constrains the outputs to be the same when the input is disturbed, assuming that models should be robust to the local perturbation \cite{sajjadi2016regularization}. Data augmentation has also been shown effective \cite{zhang2017mixup,sohn2020fixmatch}.

\subsection{Unsupervised domain adaptation (UDA)}

UDA is a popular branch of DA, which mainly focuses on distribution alignment. Recently, some deep-learning-based DA approaches have shown superior performance in extracting well aligned features \cite{wang2018deep}. A common idea is using neural networks to minimize the domain discrepancy \cite{long2015dan}. Another idea is adversarial learning \cite{ganin2015dann,long2018cdan}, which plays a two-party game between the feature extractor and the domain discriminator to obtain domain invariant feature representations. Additionally, some approaches minimize the first or second order entropy of model outputs in the target domain \cite{grandvalet2005semi,jin2020mcc}.

\subsection{Semi-supervised domain adaptation (SSDA)}
 SSDA makes use of the labeled and unlabeled target data simultaneously. MME \cite{saito2019semi} moves the prototypes to bring the source and target domain distributions together, by minimizing and maximizing the entropy alternatively. ECACL \cite{li2021semi} further aligns the prototypes of each class to enhance the discriminability, and uses FixMatch \cite{sohn2020fixmatch} to augment the unlabeled target data. Additional approaches use adversarial perturbation, e.g., BiAT \cite{jiang2020bidirectional} uses labeled source and target data to generate adversarial instances, and APE \cite{kim2020attract} adds perturbations to all target data to reduce the intra-domain discrepancy within the target domain.

\subsection{Source-free domain adaptation (SFDA)}

SFDA is a more challenging DA scenario, where the source data are unavailable during training. One common solution is to synthesize some source data. Model adaptation (MA) \cite{li2020model} produces target-style source data using generative adversarial networks (GAN) \cite{Goodfellow2014gan}. Contrastive prototype generation and adaptation (CPGA) \cite{qiu2021source} first generates feature prototypes for each class by exploring the classification boundary information of the source model, and then adopts prototype alignment. Another commonly used strategy is self-training. Source hypothesis transfer (SHOT) \cite{liang2020shot} aligns the target domain to the trained source model using information maximization and self-supervised pseudo-labeling. Neighborhood reciprocity clustering (NRC) \cite{yang2021nrc} captures the intrinsic structure of the target data using clustering hypothesis for better alignment.

\subsection{Ensemble learning}

Ensemble learning \cite{zhou2012ensemble,drwuPL2020,drwuTSK2020} constructs and aggregates multiple base learners. Boosting \cite{boosting} is a popular ensemble learning approach, which adaptively changes the sample weights according to the classification error. LogitBoost \cite{friedman2000additive} decomposes the classification problem into multiple sub-regression problems and solves them iteratively. Bagging and boosting fine-tuning (BBF) \cite{Zhao2023BBF} combines LogitBoost and extreme learning machine \cite{huang2004elm} to construct a general fine-tuning framework. Adaptive semi-supervised ensemble (ASSEMBLE) \cite{bennett2002exploiting} uses AdaBoost to train the base learners, and then assigns pseudo-labels and calculates sample weights. TrAdaBoost \cite{dai2007boosting} is an extension of boosting for DA, which views the source data misclassified by previous learners as target-unrelated, and reduces their sample weights. MultiSourceTrAdaBoost \cite{yao2010boosting} extends TrAdaBoost to multi-source DA.

Ensemble learning has also been used in deep-learning-based DA. For unsupervised DA, Nozza et al. \cite{nozza2016deep} used deep learning to extract inter-media representations, and then compared different ensemble learning strategies such as Bagging and Boosting in reducing domain distribution discrepancies. For multi-source DA, Zhou et al. \cite{zhou2021domain} proposed domain adaptation ensemble learning, which learns domain specific knowledge from each source domain and distills general knowledge from multiple domains collaboratively. For source-free DA, Ahmed et al. \cite{ahmed2022cleaning} proposed negative ensemble learning, which applies multiple data augmentations in the target domain to train multiple base learners, each of which randomly chooses complementary labels to calculate the loss and feedback. However, most existing approaches simply ensemble deep neural networks, which have high computational cost.


\begin{figure*} [htbp]     \centering
    \includegraphics[width=.9\textwidth]{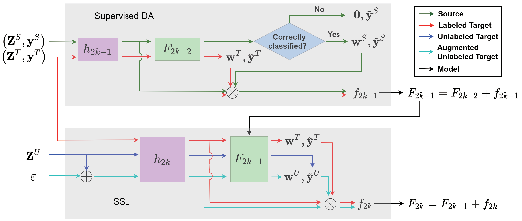}
    \caption{Structure of the $k$-th fine-tuning block. We combine labeled source data $(\bm{Z}^S,\bm{y}^S)$ and labeled target data $(\bm{Z}^T,\bm{y}^T)$ for supervised DA. The sample weights $\bm{w}$ and pseudo-labels $\tilde{\bm{y}}$ are obtained to train the base learners, as in LogitBoost. A consistency constraint is used in SSL, i.e., we add random noise $\bm{\varepsilon}$ to unlabeled target data $\bm{Z}^U$, and use pseudo-labels of $\bm{Z}^U$ to supervise the training of the augmented unlabeled target data.}
    \label{fig:fine-tuning block}
\end{figure*}

\section{Methodology}

\subsection{Problem Definition}

Consider an SSDA classification problem, which includes three subsets $\mathcal{D}^S=\{(\bm{x}_n^S,\bm{y}_n^S)\}_{n=1}^{N^S}$, $\mathcal{D}^T=\{(\bm{x}_n^T,\bm{y}_n^T)\}_{n=1}^{N^T}$ and $\mathcal{D}^U=\{\bm{x}_n^U\}_{n=1}^{N^U}$, where $\mathcal{D}^S$, $\mathcal{D}^T$ and $\mathcal{D}^U$ consist of a large number of labeled source data, a small number of labeled target data, and a large number of unlabeled target data respectively, and $\bm{y}=[y^1,...,y^J]$ is the one-hot encoding of the class label of the corresponding $\bm{x}$. Both domains share the same label space and feature space. We only consider $J>2$ in this paper; however, SS-TrBoosting can be easily modified to handle binary classification.

Deep-learning-based DA approaches often train a feature extractor $\phi$ and a classifier $f$. SS-TrBoosting takes the outputs of $\phi$ as the features, i.e., $\bm{z}=\phi(\bm{x})\in\mathbb{R}^{d\times 1}$, where $d$ is the feature dimensionality, and generates $K$ fine-tuning blocks, i.e., $2K$ base learners to enhance $f_{\mathrm{norm}}$, the normalized $f$:
\begin{align}
	F_{2K}\left(\bm{z}\right)=f_{\mathrm{norm}}(\bm{z})+\sum_{k=1}^{2K} f_k\left(\bm{z}\right). \label{eq:BF}
\end{align}
$f_{\mathrm{norm}}$ eliminates the magnitude differences between different base learners' outputs, and is computed as \cite{Zhao2023BBF}:
\begin{align}
	f_{\mathrm{norm}}(\bm{z})&=[f_{\mathrm{norm}}^1(\bm{z}),...,f_{\mathrm{norm}}^J(\bm{z})], \label{eq:norm_init}
\end{align}
where
\begin{align}
	f_{\mathrm{norm}}^j(\bm{z})&=\frac{ f^j(\bm{z})-\bar{f}(\bm{z})}{\sqrt{\sum_{i=1}^J [ f^i(\bm{z})-\bar{f}(\bm{z})]^2}},\quad j=1,...,J,\nonumber\\
	\bar{f}(\bm{z}) &= \frac{1}{J}\sum_{j=1}^J f^j(\bm{z}).
\end{align}

\subsection{SS-TrBoosting Fine-tuning Block}
In the $k$th fine-tuning block, SS-TrBoosting solves the SSDA classification problem by decomposing it into a supervised DA problem and an SSL problem, as shown in Figure~\ref{fig:fine-tuning block}:
\begin{enumerate}
    \item For supervised DA, SS-TrBoosting first generates a random nonlinear feature mapping, and then uses the labeled source data and labeled target data to generate a transferable base learner (denoted as the $(2k-1)$th base learner) to reduce the alignment bias.
    \item For SSL, SS-TrBoosting first generates another random nonlinear feature mapping, and then uses the labeled and unlabeled target data to exploit the target domain knowledge and trains the $2k$th base learner.
\end{enumerate}

\subsubsection{Random nonlinear feature mapping}
According to \cite{Zhao2023BBF}, combining boosting with extreme learning machine \cite{huang2004elm} can improve the model's generalization performance. Extreme learning machine is a single hidden layer neural network, which first randomly generates and fixes the hidden layer features, and then uses ridge regression \cite{ridge1970} to compute the output weights.

SS-TrBoosting also randomly generates the features fed into each base learner, i.e.,
\begin{align}
	h_k(\bm{z})=\delta[\mathrm{ZS}(\bm{z}^\intercal M_k,\bm{\mu}_k,\bm{\sigma}_k)],\quad k=1,...,2K, \label{eq:mapping}
\end{align}
where $\mathrm{ZS}$ is the $z$-score normalization:
\begin{align}
	\mathrm{ZS}(\bm{z}^\intercal M_k,\bm{\mu}_k,\bm{\sigma}_k) \equiv (\bm{z}^\intercal M_k-\bm{\mu}_k)./\bm{\sigma}_k,\nonumber
\end{align}
$\delta$ can be any nonlinear activation function, $M_k$ is a random matrix, and $\bm{\mu}_k$ and $\bm{\sigma}_k$ are the mean and standard deviation vectors of $Z^SM_k$, respectively, in which $Z^S= [\bm{z}^S_1,...,\bm{z}^S_{N^S}]^\intercal\in \mathbb{R}^{N^S\times d}$.

\subsubsection{Supervised DA}
We first combine the labeled source data and labeled target data to form a new dataset:
\begin{align}
	\left\{(\bm{z}_n, \bm{y}_n)\right\}_{n=1}^{N^T+N^S}=&\left\{(\bm{z}_1^T, \bm{y}_1^T),...,(\bm{z}_{N^T}^T, \bm{y}_{N^T}^T),\right.\nonumber \\
	&\left. (\bm{z}_1^S, \bm{y}_1^S),...,(\bm{z}_{N^S}^S, \bm{y}_{N^S}^S) \right\}.
\label{eq:SDA dataset}
\end{align}

Assume SS-TrBoosting has generated $2k-2$ base learners, and seeks to train $f_{2k-1}$. According to LogitBoost \cite{friedman2000additive}, we can minimize the following loss function in an additive manner:
\begin{align}
	\mathcal{L}(f_{2k-1})& =\sum_{n=1}^{N^T+N^S}  \ell \left[\bm{y}_n, F_{2k-2}\left(\bm{z}_n\right) +f_{2k-1}\left(\bm{z}_n\right)  \right], \label{eq:BF1}
\end{align}
where $\ell$ is the cross-entropy loss.

According to Taylor's Theorem, we can approximate (\ref{eq:BF1}) as:
\begin{align}
 	\mathcal{L}(f_{2k-1})  &\approx \sum_{n=1}^{N^T+N^S}  \{ \ell\left[\bm{y}_n, F_{2k-2}\left(\bm{z}_n \right)  \right] + \bm{g}_n f_{2k-1}\left(\bm{z}_n \right) \nonumber\\
 & + \frac{1}{2}\bm{q}_n f_{2k-1}^2\left(\bm{z}_n \right)  \}, \label{eq:BF2}
\end{align}
where
\begin{align}
	\bm{g}_{n}&=\frac{\partial \ell\left[\bm{y}_n, F_{2k-2}\left(\bm{z}_n \right)\right]}{\partial  F_{2k-2}\left(\bm{z}_n\right)} \label{eq:g} \\
	\bm{q}_{n}&=\frac{\partial^{2} \ell\left[\bm{y}_n,  F_{2k-2}\left(\bm{z}_n\right)\right]}{\partial  F_{2k-2}\left(\bm{z}_n\right)^{2}} \label{eq:h}
\end{align}
are the first and second order derivatives of $\ell$ w.r.t. $F_{2k-2}(\bm{z}_n)$, respectively.

Following LogitBoost \cite{friedman2000additive}, we can simplify (\ref{eq:BF2}) by removing its constant term $\sum_{n=1}^{N^T+N^S} \ell\left[y_n, F_{2k-2}\left(\bm{z}_n\right) \right]$, adding a new constant term $\sum_{n=1}^{N^T+N^S} \frac{1}{2}\frac{\bm{g}_n^2}{\bm{q}_n}$, and decomposing it into $J$ binary classification problems (e.g., $\bm{y}_n=[y_n^1, y_n^2,..., y_n^J]$, where $y_n^j$ is $+1$ or $-1$), so that:
\begin{align}
	\mathcal{L}(f_{2k-1})=\sum_{n=1}^{N^T+N^S}\sum_{j=1}^J \frac{1}{2} w_n^j \left[f_{2k-1}^j\left(\bm{z}_n\right) -\tilde{y}_n^j \right]^2, \label{eq:lossTS}
\end{align}
where
\begin{align}
	w_{n}^j&=p^j(\bm{z}_n)[1-p^j(\bm{z}_n)] \label{eq:w}
\end{align}
is the sample weight in the $j$th binary classification problem, and
\begin{align}
\tilde{y}_{n}^j&=-\frac{y_n^j-p^j(\bm{z}_n)}{w_{n}^j}\label{eq:pseudolabel}
\end{align}
is the pseudo-label, in which
\begin{align}
	p^j(\bm{z}_n)&=\mathrm{softmax}^j[F_{2k-2}(\bm{z}_n)]=\frac{e^{F_{2k-2}^j(\bm{z}_n)}}{\sum_{i=1}^J e^{F_{2k-2}^i(\bm{z}_n)}} \label{eq:output1}
\end{align}
is the estimated probability of Class $j$ for $\bm{z}_n$.

Similar to BBF \cite{Zhao2023BBF}, we clip $p^j(\bm{z}_n)$ to avoid being divided by zero in (\ref{eq:pseudolabel}):
\begin{align}
	p^j(\bm{z}_n)=\mathrm{max}\{ 0.0001, \mathrm{min} [ p^j(\bm{z}_n), 0.9999 ] \}.\label{eq:clip_p}
\end{align}

Similar to LogitBoost \cite{friedman2000additive}, we also clip the pseudo-label $\tilde{y}_{n}^j$ to increase its robustness to noise:
\begin{align}
	\tilde{y}_{n}^j=\max\left[-4,\min\left(4,\tilde{y}_n^j\right)\right].\label{eq:clip_y}
\end{align}
	
The domain alignment bias may still exist after pre-aligning the source and target distributions, which could result in negative transfer \cite{drwuNTL2022}. Inspired by TrAdaBoost \cite{dai2007boosting}, it is reasonable to assume that the source data misclassified by $F_{2k-2}$ are unrelated to the target domain, so we set their sample weights to zero:
\begin{align}
	w_{n}^j=0,\quad n>N^T,
\end{align}
if:
\begin{align}
	\mathrm{argmax}[F_{2k-2}(\bm{z}_n)]\neq \mathrm{argmax}(\bm{y}_{n}).
	\nonumber
\end{align}

\subsubsection{SSL}

For SSL, we develop a data augmentation approach to exploit the information in the unlabeled target data.

First, we use $F_{2k-1}$ to compute the pseudo-labels of the unlabeled target data:
\begin{align}
	\hat{\bm{y}}_n^U = \mathrm{OneHotCoding} \left \{ \mathrm{argmax} \left [ F_{2k-1}(\bm{z}_n^U) \right ] \right \}, \label{eq:hotlabel}
\end{align}
where $\mathrm{OneHotCoding}$ transforms the label to the one-hot encoding. Next, we add Gaussian noise
\begin{align}
    \bm{\varepsilon} \sim \mathcal{N}[\bm{0},(\xi\Sigma)^2],\label{eq:noise}
\end{align}
to the unlabeled target data, where $\xi$ controls the magnitude of the noise, $\Sigma=\mathrm{diag}[std(\bm{Z}^U)]$, in which $\bm{Z}^U= [\bm{z}^U_1,...,\bm{z}^U_{N^U}]^\intercal\in \mathbb{R}^{N^U\times d}$. We then combine them with the labeled target data to generate an augmented dataset:
\begin{align}
	\left\{(\bm{z}_n, \bm{y}_n)\right\}&_{n=1}^{N^T+N^U}= \left\{(\bm{z}_1^T, \bm{y}_1^T),...,(\bm{z}_{N^T}^T, \bm{y}_{N^T}^T),\right.\nonumber \\
	&\left. (\bm{z}_1^U+\bm{\varepsilon}_1, \hat{\bm{y}}_1^U),...,(\bm{z}_{N^U}^U+\bm{\varepsilon}_{N^U}, \hat{\bm{y}}_{N^U}^U) \right\}.
\label{eq:SSL dataset}
\end{align}

Then, we minimize the following loss to train $f_{2k}$:
\begin{align}
	\mathcal{L}(f_{2k})=\sum_{n=1}^{N^T+N^U}\sum_{j=1}^J \frac{1}{2} w_n^j \left[f_{2k}^j\left(\bm{z}_n\right) -\tilde{y}_n^j \right]^2,\label{eq:lossTU}
\end{align}
where $w_n^j$ and $\tilde{y}_n^j$ are computed using (\ref{eq:w}) and (\ref{eq:pseudolabel}), respectively, and $p^j(\bm{z}_n)$ in (\ref{eq:w}) and (\ref{eq:pseudolabel}) is computed as:
\begin{align}
	p^j(\bm{z}_n)&=\nonumber\\
	&\begin{cases}
		\mathrm{softmax}^j[F_{2k-1}(\bm{z}_n^T)],&n\leq N^T\\
		\mathrm{softmax}^j[F_{2k-1}^{\mathrm{noise}}(\bm{z}_{n-N^T}^U,\bm{\varepsilon}_{n-N^T})],&n> N^T
	\end{cases},
\label{eq:output2}
\end{align}
in which
\begin{align}
	F_{2k-1}^{\mathrm{noise}}&(\bm{z},\bm{\varepsilon})=\nonumber\\
	&\begin{cases}
		f_{\mathrm{norm}}(\bm{z}+\bm{\varepsilon}),&k=1\\
		F_{2k-2}(\bm{z})+f_{2k-1}\left(\bm{z}+\bm{\varepsilon}\right),&k>1
	\end{cases}.
\end{align}
Using $F^{\mathrm{noise}}(\cdot)$ to get the noisy data's outputs can speed up the training process.

Note that we also use (\ref{eq:clip_p}) and (\ref{eq:clip_y}) to clip the output probability $p^j(\bm{z}_n)$ and pseudo-label $\tilde{y}_n^j$, respectively.

\subsubsection{Implementation details}

The balanced sampling (BS) trick in \cite{Zhao2023BBF} can reduce the computational cost and enable the model to handle class-imbalance problems. To balance the number of labeled target data and labeled source data, or the number of labeled and unlabeled target data, in each batch, we first use BS to sample two batches of points from them, and then merge them to train a base learner.

Additionally, we use the Norm trick in \cite{Zhao2023BBF} to ensure the outputs of the initial model and  the base learners have the same order of magnitude.

The pseudo-code of SS-TrBoosting is given in Algorithm~\ref{Alg:SS-TrBoosting}, where the pseudo-code of BS and  Norm is given in Algorithms~\ref{Alg:BS} and \ref{Alg:Norm}, respectively.

\begin{algorithm*}[htbp]
	\caption{SS-TrBoosting.}\label{Alg:SS-TrBoosting}
	\KwIn{$\{(\bm{z}_n^S,\bm{y}_n^S)\}_{n=1}^{N^S}$, $N^S$ labeled source samples; $\{(\bm{z}_n^T,\bm{y}_n^T)\}_{n=1}^{N^T}$, $N^T$ labeled target samples; $\{\bm{z}_n^U\}_{n=1}^{N^U}$, $N^U$ unlabeled target samples; $K$, the number of fine-tuning blocks; $J$, the number of classes ($J>2$); $f_c$, the initial model; $ns$, the node size; $lr$, the learning rate; $bs$, the batch size; $\xi$, magnitude of the noise.}
	
	\KwOut{$F$, an ensemble classifier.}
	\vspace*{2mm}
	Construct $\bm{Z}^S= [\bm{z}^S_1,...,\bm{z}^S_{N^S}]^\intercal\in \mathbb{R}^{N^S\times d}$;\quad  // $\bm{z}\in \mathbb{R}^{d\times 1}$
	Compute the normalized model $f_{\mathrm{norm}}(\bm{z})$ using (\ref{eq:norm_init})\;
	Initialize $F(\bm{z})=f_{\mathrm{norm}}(\bm{z})$\;
	\For{$k=1:2K$}{
		Randomly generate a matrix $\bm{W}_{h}\in \mathbb{R}^{d\times ns}$\;
		Construct $h_k(\bm{z})=\delta\{\mathrm{ZS}[\bm{z}^\intercal\bm{W}_{h}, mean(\bm{Z}^S\bm{W}_{h}),std(\bm{Z}^S\bm{W}_{h})]\}$\;
		\For{$j=1:J$}{
			\uIf{$\mathrm{mod}(k,2)==1$ }{
				Generate the DA training set $\left\{(\bm{z}_n, \bm{y}_n)\right\}_{n=1}^{N^T+N^S}$ using (\ref{eq:SDA dataset})\;
				\For{$n=1:N^T+N^S$}{
					Compute $\bm{p}^j(\bm{z}_n)$ using (\ref{eq:output1})\;
					Clip $\bm{p}^j(\bm{x}_n)$ using  (\ref{eq:clip_p})\;
					Compute pseudo-label $\tilde{\bm{y}}_n^j$ using (\ref{eq:pseudolabel})\;
					Clip $\tilde{\bm{y}}_n^j$ using  (\ref{eq:clip_y})\;
					Compute sample weight $\bm{w}_n^j$ using (\ref{eq:w})\;
					\If{$n>N^T$ $\mathrm{and}$ $\mathrm{argmax}[F_{k-1}(\bm{z}_n)]\neq \mathrm{argmax}(\bm{y}_n)$}{
						$\bm{w}_n^j=0$\;
					}
				}
				Generate the batch index set $I_{\mathrm{batch}}^T=\mathrm{BS}(\{(\bm{y}_n,\bm{w}_n)\}_{n=1}^{N^T},~J,~bs,~j)$\;
				Generate the batch index set $I_{\mathrm{batch}}^S=\mathrm{BS}(\{(\bm{y}_n,\bm{w}_n)\}_{n=N^T+1}^{N^T+N^S},~J,~bs,~j)$\;
				$I_{\mathrm{batch}}^j=I_{\mathrm{batch}}^T\cup I_{\mathrm{batch}}^S$.
			}
			\Else{
				Generate the noise $\bm{\varepsilon}$ using (\ref{eq:noise})\;
				Generate the pseudo-labels of the unlabeled target data using (\ref{eq:hotlabel})\;
				Generate the SSL training set $\left\{(\bm{z}_n, \bm{y}_n)\right\}_{n=1}^{N^T+N^U}$using (\ref{eq:SSL dataset})\;
				\For{$n=1:N^T+N^U$}{
					Compute $\bm{p}^j(\bm{z}_n)$ using (\ref{eq:output2})\;
					Clip $\bm{p}^j(\bm{x}_n)$ using  (\ref{eq:clip_p})\;
					Compute pseudo-label $\tilde{\bm{y}}_n^j$ using (\ref{eq:pseudolabel})\;
					Clip $\tilde{\bm{y}}_n^j$ using  (\ref{eq:clip_y})\;
					Compute sample weight $\bm{w}_n^j$ using (\ref{eq:w})\;
				}
				Generate the batch index $I_{\mathrm{batch}}^T=\mathrm{BS}(\{(\bm{y}_n,\bm{w}_n)\}_{n=1}^{N^T},~J,~bs,~j)$\;
				Generate the batch index $I_{\mathrm{batch}}^U=\mathrm{BS}(\{(\bm{y}_n,\bm{w}_n)\}_{n=N^T+1}^{N^T+N^U},~J,~bs,~j)$\;
				$I_{\mathrm{batch}}^j=I_{\mathrm{batch}}^T\cup I_{\mathrm{batch}}^U$.
			}				
			
			Use ridge regression to train $f_k^j(\bm{z})$ on $\{(h_k(\bm{z}_n),\tilde{\bm{y}}_n^j)\}_{n\in I_{\mathrm{batch}}^j}$\;
		}
		
		Update $F(\bm{z}) \leftarrow F(\bm{z}) + lr \times \mathrm{Norm}\{f_k[h_k(\bm{z}),J]\}$\;
	}
\end{algorithm*}

\begin{algorithm}[htpb]
	\caption{Balanced sampling (BS) for $J$-class classification \cite{Zhao2023BBF}.}\label{Alg:BS}
	\KwIn{$\{(\bm{w}_n,\bm{y}_n)\}_{n=1}^N$, $N$ sample weights and labels\;
		\hspace*{10mm} $J$, the number of classes ($J>2$)\;
		\hspace*{10mm} $bs$, the batch size\;
		\hspace*{10mm} $\hat{j}$, the positive class, default 1.}
	\KwOut{$I_{\mathrm{batch}}$, the index set of the data in one batch.}
	\vspace*{2mm}
	Let $\mathrm{WS}(I,bs)$ denote weighted sampling with replacement of $bs$ samples from $I$\;
	Let $\mathrm{DS}(I,bs)$ denote down-sampling without replacement of $bs$ samples from $I$\;

	Initialize $I_{\mathrm{neg}}=\{\}$\;
	\For{$j=1:J$}{	// $\bm{y}_n\in \mathbb{R}^{J\times2},~n=1,...,N$\;
		\uIf{$j ==\hat{j}$ }{
			$\displaystyle I_{\mathrm{pos}}=\mathrm{WS}(\{(\bm{w}_n,\bm{y}_n)|~\bm{y}_n^j==1, n\in[1,N]\},bs)$\;
		}
		\Else{
			$\displaystyle I'= \mathrm{WS}(\{(\bm{w}_n,\bm{y}_n)|~\bm{y}_n^j==1, n\in[1,N]\},bs)$\;
			$I_{\mathrm{neg}}=I_{\mathrm{neg}}\cup I'$\;
		}
	}
	$I_{\mathrm{neg}}=\mathrm{DS}(I_{\mathrm{neg}},bs)$\;

	$I_{\mathrm{batch}}=I_{\mathrm{pos}}\cup I_{\mathrm{neg}}$.
\end{algorithm}


\begin{algorithm}[htpb]
	\caption{$\mathrm{Norm}$ for $J$-class classification \cite{Zhao2023BBF}.}\label{Alg:Norm}
	\KwIn{$f$, original output of the base learner\;
		\hspace*{10mm} $J$, the number of classes ($J> 2$).}
	\KwOut{$f_{\mathrm{norm}}$, the normalized output of the base learner.}
	\vspace*{2mm}
	
	Denote $f=[f^1,...,f^J]^T$; \quad 		// $f\in \mathbb{R}^{J\times1}$\;
	$f_m=\frac{1}{J}\sum_{j=1}^Jf^j$\;
	\For{$j=1:J$}{
		$f^j=\frac{J-1}{J}(f^j-f_m)$\;
		$f^j=\max\left[-4,\min(4,f^j)\right]$\;
	}
	$f_{\mathrm{norm}}^j = \frac{f^j}{\sqrt{ \sum_{i=1}^{J} (f^i)^2}} $, \quad $j=1,...,J$\;
	Construct $f_{\mathrm{norm}}=[f_{\mathrm{norm}}^1,...,f_{\mathrm{norm}}^J]^T$\;
\end{algorithm}

\subsubsection{Summary}

Compared with traditional boosting approaches, SS-TrBoosting has three distinct characteristics:
\begin{enumerate}
 \item Random nonlinear feature mapping in both supervised DA and SSL, which introduces randomness into the training process.
 \item Data re-weighting in supervised DA, which aligns the class-conditional probability distribution between domains by forcing later base learners to pay more attention to the more difficult target samples and the source samples more relevant to the target domain. SS-TrBoosting views source data misclassified by previous learners as target-irrelevant, and sets their sample weights to zero, which helps reduce the domain alignment bias.
 \item Data augmentation in SSL, which generates pseudo-labels and adds Gaussian noise to the extracted features of the unlabeled target data.
\end{enumerate}
The effectiveness of the three characteristics is demonstrated in the next section.

\subsection{Extension to SS-SFDA}

SS-TrBoosting cannot be used when the source data are unavailable. This subsection proposes a virtual source domain generation approach to extend SS-TrBoosting to SS-SFDA, which can be used to enhance the performance of a pre-trained SFDA model. A well trained classifier contains information about the training data distribution  \cite{nayak2021mining}, so synthetic source data can be generated from it.

Our proposed source data generation approach consists of three steps:
\begin{enumerate}
\item Randomly generate $N^L$ label vectors for each class. Let $\hat{\bm{y}}_{c,i}=[\hat{y}_{c,i}^1,\hat{y}_{c,i}^2,...,\hat{y}_{c,i}^J]^{\intercal}$ be the $i$-th ($i=1,2,...,N^L$) generated label vector for Class $c$ ($c=1,2,...,J$). To increase the diversity of the generated label vectors, we set $\hat{y}_{c,i}^j$ as:
\begin{align}
    \hat{y}_{c,i}^j=
    \begin{cases}
        \alpha_i, &j=c \\
        1-\alpha_i, &j=r \\
        0, &\mbox{otherwise}
    \end{cases},
\end{align}
where $r\neq c$ is a randomly selected class, and $\alpha_i=\max \left [\beta_i, 1-\beta_i \right ]$, in which $\beta_i\sim \mathrm{Beta}(a,b)$ is a random number from a Beta distribution ($a$ and $b$ are hyper-parameters). We set $a=0.75$ and $b=0.75$ in our experiments, i.e., $\beta \sim \mathrm{Beta}(0.75,0.75)$.

\item Synthesize the source data by using the generated label vectors. Given the source features $\bm{Z}^S \in \mathbb{R}^{d \times JN^L}$ and the generated $\hat{\bm{Y}}^S=[\hat{\bm{y}}_{1,1},...,\hat{\bm{y}}_{1,N^L},\hat{\bm{y}}_{2,1},...,
    \hat{\bm{y}}_{2,N^L},...,\hat{\bm{y}}_{J,1},...,\hat{\bm{y}}_{J,N^L}]\in \mathbb{R}^{J\times JN^L}$, we can train a linear model with parameter $\bm{\theta} \in \mathbb{R}^{J\times d}$ by solving $\bm{\theta}\bm{Z}^S=\bm{Y}^S$. $\bm{Z}^S$ can then be estimated by:
\begin{align}
    \hat{\bm{Z}}^S = \bm{\theta}^{\dagger} \hat{\bm{Y}}^S,
\end{align}
where $\bm{\theta}^{\dagger}$ is the pseudo-inverse of $\bm{\theta}$. In practice, we take the parameters of the last linear layer of the pre-trained SFDA network as $\bm{\theta}$, and use ridge regression \cite{ridge1970} to get $\hat{\bm{Z}}^S$.

\item Align the mean and standard deviation of the synthetic data and target data:
\begin{align}
    \hat{\bm{Z}}^{ST} = \frac{\left ( \hat{\bm{Z}}^S-\bm{\mu}^S \right )\bm{\sigma}^T }{\bm{\sigma}^S}  + \bm{\mu}^T,
\end{align}
where $\bm{\mu}^S$ ($\bm{\mu}^T$) and $\bm{\sigma}^S$ ($\bm{\sigma}^T$) are mean and standard deviation of $\hat{\bm{Z}}^S$ ($\hat{\bm{Z}}^T$), respectively, and $\hat{\bm{Z}}^{ST}$ is the generated data.

\end{enumerate}

\section{Experiments}\label{sec:Experimet}

Extensive experiments were carried out to verify the effectiveness, flexibility and robustness of our proposed SS-TrBoosting.

\subsection{Datasets}

DomainNet \cite{peng2019moment} is a large-scale dataset, which contains 345 classes and six domains. We picked four domains (R: Real, C: Clipart, P: Painting, S: Sketch) to construct 12 SSDA scenarios.

Office-Home \cite{venkateswara2017deep} is a medium-scale dataset, which contains 65 classes and four domains (R: Real, C: Clipart, A: Art, P: Product). We constructed 12 SSDA scenarios from it.

Office-31 \cite{saenko2010adapting} is a relatively small dataset, which contains 31 classes with three domains (A: Amazon, W: Webcam, D: DSLR). We constructed six SSDA scenarios from it.

\subsection{Baselines}

For SSDA, simple transfer (ST) is a baseline trained with labeled data in both domains. DAN \cite{long2015dan}, DANN \cite{ganin2015dann}, CDAN \cite{long2018cdan}, FixBi \cite{na2021fixbi}, DNA \cite{chen2023domain}, and MCC \cite{jin2020mcc} are UDA baselines. ENT \cite{grandvalet2005semi}, MME \cite{saito2019semi}, UODA \cite{qin2021contradictory} and APE \cite{kim2020attract} are SSDA baselines. We modified all UDA baselines to use the labeled target data, by combining them with the labeled source data.

For SS-SFDA, source only (SourceOnly) means directly using the model trained in the source domain, without any adaptation. SHOT \cite{liang2020shot} and NRC \cite{yang2021nrc} are SFDA baselines. We modified all three baselines to use the labeled target data, by adding a supervised training loss.

\subsection{Implementation Details}

For all UDA and SSDA baselines, we used ResNet-50 pre-trained on ImageNet as the base network, and added two linear layers after it: 1) a bottleneck layer that maps the outputs to 2,048 dimensions, whose outputs were then used as the input features of SS-TrBoosting; 2) a linear classification layer, which was used as the initial model in SS-TrBoosting. For SFDA baselines, we modified the output dimensionality of the bottleneck layer from 2,048 to 256 and kept all other settings unchanged. When training the deep networks, we adopted stochastic gradient descent with a momentum of 0.9, an initial learning rate of 0.01, and weight decay of 0.0005. The random seed was 2021. Other settings followed the original papers that proposed those baselines, respectively.

For SS-TrBoosting in SSDA, the number of fine-tuning blocks, batch size, and $\xi$ were set to 100, 64 and 1.0, respectively. In each iteration, SS-TrBoosting randomly mapped the input features into 100 dimensions, i.e., $M_K\in \mathbb{R}^{d\times 100}$. $N^L=100$ was used in generating $Z^S$ for SS-SFDA, i.e., 100 samples were generated for each class.

For Office-31 and Office-Home, the results were averaged from three repeats. For DomainNet, we only ran the experiments once due to the excessive computational cost.

All source code was implemented in PyTorch, and ran on a server with a NVIDIA RTX 3090 GPU and 112GB RAM.

\subsection{SSDA Results}

\subsubsection{Overview}

Tables~\ref{tab:domainNet}-\ref{tab:office-31} show the results of SS-TrBoosting for SSDA on the three datasets, respectively. ``$n$-shot" means only $n$ samples per class in the target domain were labeled.

SS-TrBoosting improved the performances of almost all baselines, demonstrating that SS-TrBoosting is compatible with various DA approaches. Generally, SSDA baselines outperformed UDA baselines, due to their abilities to use the labeled target data.

Moreover, the baselines in the 3-shot setting outperformed their counterparts in 1-shot setting by about 3.6$\%$ on average, demonstrating that few labeled target data can noticeably improve the performance of the DA approaches.

\begin{table*}[ht]
	\caption{Accuracies ($\%$) on DomainNet for 1-shot and 3-shot settings for SSDA, when SS-TrBoosting was used to fine-tune.}
	\renewcommand\arraystretch{1.1}  \small  \centering \setlength{\tabcolsep}{1.8mm}
	\begin{tabular}{c|c|cccccccccccc|c}
		\toprule
		\multicolumn{2}{c|}{Algorithm} & C$\rightarrow$P & C$\rightarrow$R & C$\rightarrow$S & P$\rightarrow$C & P$\rightarrow$R & P$\rightarrow$S & R$\rightarrow$C & R$\rightarrow$P & R$\rightarrow$S & S$\rightarrow$C & S$\rightarrow$P & S$\rightarrow$R & Avg \\
		\hline
		\multicolumn{15}{c}{\textbf{1-shot}} \\
		\hline
		&ST    & 39.98 & 58.33 & 42.96 & 45.84 & 61.41 & 39.80 & 51.97 & 49.77 & 41.00 & 53.89 & 42.62 & 55.31 & 48.57 \\
		&SS-TrBoosting & 43.30 & 62.39 & 45.40 & 49.94 & 64.98 & 42.26 & 54.14 & 51.23 & 42.60 & 56.82 & 45.67 & 60.63 & \textbf{51.61} \\ \hline
		\multirow{8}{*}{UDA}
		&DAN   & 39.61 & 57.48 & 42.82 & 44.11 & 60.39 & 39.77 & 48.78 & 48.94 & 39.35 & 52.96 & 42.40 & 56.89 & 47.79 \\
		&SS-TrBoosting & 43.10 & 61.94 & 45.19 & 48.26 & 64.13 & 41.79 & 50.22 & 49.91 & 40.83 & 55.82 & 45.34 & 60.73 & \textbf{50.61} \\\cline{2-15}
		&DANN  & 40.13 & 57.47 & 43.83 & 46.15 & 60.97 & 40.53 & 52.64 & 49.41 & 41.46 & 55.06 & 44.83 & 56.97 & 49.12 \\
		&SS-TrBoosting & 44.67 & 62.04 & 46.63 & 50.15 & 63.97 & 42.42 & 54.60 & 51.30 & 42.96 & 58.29 & 47.47 & 60.81 & \textbf{52.11} \\\cline{2-15}
		&CDAN  & 44.68 & 61.52 & 47.51 & 50.57 & 63.04 & 45.41 & 58.83 & 52.72 & 45.32 & 61.30 & 49.10 & 60.04 & 53.34 \\
		&SS-TrBoosting & 45.55 & 63.10 & 48.93 & 51.85 & 64.32 & 45.73 & 59.00 & 52.83 & 45.66 & 62.11 & 50.14 & 61.75 & \textbf{54.25} \\\cline{2-15}
        &DNA   & 42.23  & 58.70 & 45.97 & 47.03 & 60.16 & 42.25 & 53.28 & 50.12 & 42.48 & 56.23 & 45.17 & 55.87 & 49.96 \\
		&SS-TrBoosting & 44.72 & 61.49 & 48.34 & 48.86 & 62.47 & 43.81 & 54.29 & 51.43 & 43.65 & 58.86 & 46.99 & 58.72 & \textbf{51.97} \\\cline{2-15}
		&MCC   & 45.41 & 61.36 & 49.00 & 50.42 & 61.69 & 45.68 & 56.54 & 52.64 & 45.48 & 60.10 & 48.48 & 59.60 & 53.03 \\
		&SS-TrBoosting & 46.55 & 63.13 & 49.53 & 51.54 & 63.47 & 46.16 & 55.96 & 52.23 & 45.18 & 61.21 & 48.89 & 61.26 & \textbf{53.76} \\ \hline
		\multirow{6}{*}{SSDA}
		&ENT   & 39.48 & 62.47 & 44.62 & 46.20 & 63.34 & 37.03 & 50.29 & 51.08 & 36.91 & 57.75 & 46.52 & 59.05 & 49.56 \\
		&SS-TrBoosting & 43.35 & 63.89 & 46.96 & 48.92 & 64.69 & 38.98 & 53.19 & 53.02 & 40.12 & 59.40 & 48.56 & 61.55 & \textbf{51.89} \\\cline{2-15}
		&MME   & 43.56 & 59.70 & 48.10 & 49.15 & 59.40 & 44.36 & 58.06 & 51.41 & 45.83 & 59.18 & 47.59 & 58.19 & 52.04 \\
		&SS-TrBoosting & 45.39 & 61.10 & 49.42 & 50.08 & 61.90 & 45.12 & 58.13 & 51.44 & 46.14 & 60.25 & 48.58 & 60.49 & \textbf{53.17} \\\cline{2-15}
        &UODA  & 41.61 & 58.68 & 48.00 & 49.91 & 60.00 & 44.49 & 56.95 & 50.62 & 42.76 & 59.15 & 47.08 & 56.69 & 51.33 \\
        &SS-TrBoosting & 44.62 & 60.72 & 48.78 & 50.77 & 62.19 & 44.96 & 56.78 & 50.65 & 43.59 & 60.05 & 46.84 & 58.57 & \textbf{52.38}
\\\cline{2-15}
		&APE   & 35.03 & 57.45 & 41.60 & 47.35 & 58.82 & 42.40 & 53.71 & 49.82 & 42.31 & 58.77 & 45.70 & 55.92 & 49.07 \\
		&SS-TrBoosting & 38.51 & 58.66 & 44.75 & 48.55 & 60.71 & 43.11 & 55.40 & 51.46 & 43.17 & 59.62 & 46.74 & 56.90 & \textbf{50.63} \\\hline
		\multicolumn{15}{c}{\textbf{3-shot}} \\\hline
		&ST    & 44.30 & 62.14 & 46.82 & 51.30 & 64.43 & 44.31 & 55.45 & 51.98 & 45.04 & 57.53 & 46.34 & 60.93 & 52.55 \\
		&SS-TrBoosting & 47.68 & 65.83 & 48.71 & 55.11 & 67.09 & 46.23 & 57.69 & 53.67 & 46.97 & 60.19 & 49.61 & 64.91 & \textbf{55.31} \\\hline
		\multirow{8}{*}{UDA}
		&DAN   & 43.46 & 61.54 & 45.49 & 50.21 & 63.43 & 43.56 & 52.54 & 50.22 & 43.00 & 56.80 & 46.17 & 59.36 & 51.31 \\
		&SS-TrBoosting & 47.14 & 65.37 & 47.72 & 54.04 & 66.52 & 45.39 & 54.41 & 52.13 & 44.62 & 59.12 & 49.10 & 63.64 & \textbf{54.10} \\\cline{2-15}
		&DANN  & 44.57 & 61.67 & 46.34 & 50.69 & 64.04 & 43.56 & 54.86 & 51.53 & 44.23 & 58.06 & 47.81 & 60.35 & 52.31 \\
		&SS-TrBoosting & 48.06 & 65.09 & 48.94 & 54.73 & 66.22 & 45.71 & 57.19 & 53.26 & 46.29 & 61.08 & 50.56 & 64.52 & \textbf{55.14} \\\cline{2-15}
		&CDAN  & 50.15 & 65.34 & 50.38 & 57.64 & 67.13 & 50.03 & 63.53 & 56.98 & 51.41 & 64.56 & 53.41 & 64.50 & 57.92 \\
		&SS-TrBoosting & 50.69 & 66.37 & 50.98 & 58.43 & 67.40 & 50.49 & 63.53 & 57.13 & 51.74 & 65.29 & 53.80 & 65.56 & \textbf{58.45} \\\cline{2-15}
        &DNA   & 46.08  & 61.92  & 48.53  & 52.03  & 63.19  & 45.08  & 56.96  & 52.04  & 46.31  & 58.32  & 47.82  & 59.65  & 53.16  \\
		&SS-TrBoosting & 48.01  & 64.11  & 50.26  & 53.57  & 65.57  & 46.54  & 57.35  & 52.81  & 46.92  & 60.76  & 49.25  & 61.90 & \textbf{54.75} \\\cline{2-15}
		&MCC   & 50.18 & 65.11 & 50.80 & 55.88 & 65.41 & 48.49 & 61.54 & 54.71 & 49.05 & 63.08 & 52.28 & 62.60 & 56.59 \\
		&SS-TrBoosting & 50.26 & 65.87 & 51.75 & 57.07 & 65.98 & 48.69 & 60.31 & 54.48 & 49.19 & 63.40 & 51.93 & 64.15 & \textbf{56.92} \\\hline
		\multirow{6}{*}{SSDA}
		&ENT   & 50.64 & 67.40 & 50.48 & 56.17 & 68.24 & 47.60 & 60.27 & 56.48 & 48.42 & 63.31 & 53.06 & 65.85 & 57.33 \\
		&SS-TrBoosting & 52.11 & 68.01 & 51.28 & 57.34 & 68.37 & 48.29 & 61.09 & 57.35 & 49.52 & 64.04 & 54.29 & 67.08 & \textbf{58.23} \\\cline{2-15}
		&MME   & 48.54 & 64.27 & 51.10 & 55.63 & 64.77 & 48.46 & 62.36 & 53.03 & 50.05 & 63.21 & 51.35 & 62.72 & 56.29 \\
		&SS-TrBoosting & 49.88 & 65.38 & 51.41 & 55.86 & 65.97 & 49.01 & 62.32 & 53.14 & 49.95 & 63.96 & 51.54 & 63.94 & \textbf{56.86} \\\cline{2-15}
        &UODA  & 47.10 & 63.17 & 50.18 & 55.13 & 64.26 & 47.46 & 61.04 & 53.65 & 48.67 & 62.89 & 50.13 & 61.02 & 55.39 \\
        &SS-TrBoosting & 48.77 & 65.11 & 51.34 & 56.00 & 65.41 & 48.27 & 61.08 & 53.61 & 49.27 & 63.72 & 50.37 & 62.71 & \textbf{56.30}
\\\cline{2-15}
		&APE   & 42.33 & 63.45 & 49.59 & 53.39 & 63.26 & 46.73 & 60.10 & 54.01 & 48.44 & 61.66 & 49.54 & 61.29 & 54.48 \\
		&SS-TrBoosting & 44.49 & 63.81 & 50.19 & 54.25 & 64.09 & 47.56 & 60.59 & 54.27 & 49.18 & 62.16 & 50.15 & 61.99 & \textbf{55.23} \\
		\bottomrule
	\end{tabular}%
	\label{tab:domainNet}%
\end{table*}%

\begin{table*}[ht]
	\caption{Accuracies ($\%$) on Office-Home for 1-shot and 3-shot settings for SSDA, when SS-TrBoosting was used to fine-tune.}
	\renewcommand\arraystretch{1.1}  \small  \centering \setlength{\tabcolsep}{1.5mm}
	\begin{tabular}{c|c|cccccccccccc|c}
		\toprule
		\multicolumn{2}{c|}{Algorithm} & A$\rightarrow$C & A$\rightarrow$P & A$\rightarrow$R & C$\rightarrow$A & C$\rightarrow$P & C$\rightarrow$R & P$\rightarrow$A & P$\rightarrow$C & P$\rightarrow$R & R$\rightarrow$A & R$\rightarrow$C & R$\rightarrow$P & Avg (std) \\
		\hline
		\multicolumn{15}{c}{\textbf{1-shot}} \\
		\hline
		&ST    & 49.25 & 72.12 & 75.91 & 56.31 & 68.66 & 68.61 & 55.60 & 45.81 & 73.97 & 65.18 & 50.55 & 79.42 & 63.45 (0.31) \\
		&SS-TrBoosting & 52.65 & 77.35 & 78.70 & 60.41 & 74.46 & 73.63 & 59.14 & 48.76 & 77.80 & 66.82 & 53.50 & 81.98 & \textbf{67.10 (0.22)} \\ \hline
		\multirow{8}{*}{UDA}
		&DAN   & 48.49 & 70.16 & 74.64 & 53.22 & 66.74 & 66.64 & 53.12 & 43.83 & 72.70 & 63.46 & 48.57 & 77.79 & 61.61 (0.01) \\
		&SS-TrBoosting & 51.94 & 75.32 & 78.18 & 58.16 & 72.47 & 72.28 & 57.08 & 47.26 & 76.51 & 65.31 & 51.29 & 80.38 & \textbf{65.52 (0.16)} \\\cline{2-15}
		&DANN  & 48.33 & 69.68 & 74.11 & 58.43 & 67.97 & 67.81 & 56.58 & 45.52 & 74.11 & 65.27 & 50.81 & 79.11 & 63.14 (0.17) \\
		&SS-TrBoosting & 51.62 & 74.27 & 77.92 & 62.59 & 73.01 & 73.87 & 60.08 & 49.21 & 78.43 & 67.71 & 53.36 & 81.53 & \textbf{66.97 (0.13)} \\\cline{2-15}
		&CDAN  & 55.36 & 73.46 & 76.66 & 62.90 & 74.26 & 73.65 & 63.15 & 53.08 & 78.39 & 71.30 & 58.43 & 81.95 & 68.55 (0.42) \\
		&SS-TrBoosting & 55.71 & 75.29 & 78.17 & 63.35 & 74.92 & 74.15 & 63.56 & 53.81 & 79.85 & 71.45 & 58.63 & 82.75 & \textbf{69.30 (0.44)} \\\cline{2-15}
        &FixBi & 51.09 & 75.07 & 77.25 & 62.94 & 77.15 & 74.42 & 63.11 & 54.17 & 78.91 & 70.18 & 57.29 & 83.18 & 68.73 (0.16) \\
        &SS-TrBoosting & 51.09 & 75.09 & 77.29 & 62.94 & 77.17 & 74.42 & 63.11 & 54.19 & 78.91 & 70.18 & 57.30 & 83.19 & \textbf{68.74 (0.16)}
\\\cline{2-15}
        &DNA & 51.95  & 72.49  & 75.98  & 60.50  & 69.50  & 68.74  & 59.75  & 50.29  & 76.26  & 68.22  & 54.77  & 81.31  & 65.81 (0.72) \\
        &SS-TrBoosting & 54.17 & 76.37  & 79.38  & 64.13  & 74.70  & 74.49  & 63.56  & 52.45  & 79.29  & 70.51  & 56.32  & 83.13 & \textbf{69.04 (0.47)}
\\\cline{2-15}
		&MCC   & 57.52 & 79.09 & 80.59 & 65.42 & 78.44 & 78.47 & 64.20 & 55.29 & 81.09 & 70.66 & 59.13 & 83.57 & 71.12 (0.37) \\
		&SS-TrBoosting & 57.70 & 79.28 & 80.80 & 65.45 & 78.55 & 78.57 & 64.21 & 55.36 & 81.14 & 70.72 & 59.21 & 83.71 & \textbf{71.22 (0.37)} \\ \hline
		\multirow{6}{*}{SSDA}
		&ENT   & 51.41 & 74.31 & 77.91 & 63.12 & 75.30 & 74.41 & 61.70 & 50.64 & 78.67 & 70.31 & 54.32 & 81.97 & 67.84 (0.25) \\
		&SS-TrBoosting & 53.33 & 76.44 & 79.48 & 64.11 & 76.44 & 76.41 & 62.40 & 51.86 & 79.59 & 70.70 & 55.78 & 82.64 & \textbf{69.10 (0.20)} \\\cline{2-15}
		&MME   & 55.78 & 73.13 & 76.32 & 62.36 & 73.58 & 72.10 & 60.41 & 56.10 & 78.20 & 70.12 & 59.15 & 83.20 & 68.37 (0.37) \\
		&SS-TrBoosting & 56.50 & 74.74 & 77.60 & 62.90 & 75.17 & 72.55 & 60.53 & 56.46 & 79.16 & 70.24 & 59.34 & 83.54 & \textbf{69.06 (0.33)} \\\cline{2-15}
        &UODA  & 54.40 & 71.20 & 75.02 & 59.72 & 71.48 & 69.69 & 59.98 & 55.66 & 76.21 & 68.60 & 59.22 & 82.27 & 66.95 (0.12) \\
        &SS-TrBoosting & 57.45 & 76.02 & 78.04 & 61.12 & 74.56 & 72.48 & 61.91 & 56.20 & 78.70 & 70.24 & 59.79 & 83.25 & \textbf{69.15 (0.13)}
\\\cline{2-15}
		&APE   & 56.43 & 76.69 & 78.55 & 66.65 & 76.60 & 75.77 & 66.22 & 56.34 & 80.29 & 73.62 & 60.83 & 83.46 & 70.95 (0.36) \\
		&SS-TrBoosting & 57.29 & 77.35 & 79.79 & 67.53 & 77.37 & 76.79 & 66.71 & 56.98 & 80.68 & 73.62 & 61.13 & 83.84 & \textbf{71.59 (0.36)} \\
		\hline
		\multicolumn{15}{c}{\textbf{3-shot}} \\
		\hline
		&ST    & 54.40 & 76.60 & 77.85 & 61.13 & 74.10 & 71.71 & 60.99 & 53.17 & 76.39 & 68.06 & 57.01 & 81.83 & 67.77 (0.27) \\
		&SS-TrBoosting & 57.39 & 80.08 & 80.38 & 64.77 & 78.90 & 75.63 & 63.87 & 56.00 & 79.45 & 70.28 & 59.28 & 83.62 & \textbf{70.80 (0.18)} \\\hline
		\multirow{8}{*}{UDA}
		&DAN   & 52.69 & 74.93 & 76.17 & 58.90 & 71.86 & 70.29 & 56.71 & 49.45 & 74.61 & 65.41 & 53.75 & 79.61 & 65.36 (0.09) \\
		&SS-TrBoosting & 55.73 & 78.88 & 79.18 & 62.63 & 77.45 & 74.91 & 60.56 & 53.02 & 78.26 & 67.58 & 56.39 & 81.89 & \textbf{68.87 (0.08)} \\\cline{2-15}
		&DANN  & 53.56 & 75.97 & 76.44 & 62.11 & 74.75 & 72.06 & 61.42 & 52.84 & 76.36 & 68.19 & 56.25 & 81.27 & 67.60 (0.20) \\
		&SS-TrBoosting & 56.54 & 79.36 & 78.93 & 66.19 & 78.62 & 76.33 & 64.55 & 55.99 & 79.75 & 70.61 & 58.91 & 83.46 & \textbf{70.77 (0.24)} \\\cline{2-15}
		&CDAN  & 59.66 & 78.19 & 78.87 & 66.55 & 77.84 & 76.61 & 65.44 & 61.23 & 80.93 & 73.25 & 63.78 & 84.61 & 72.25 (0.32) \\
		&SS-TrBoosting & 60.45 & 79.01 & 79.59 & 66.76 & 78.66 & 77.40 & 66.23 & 61.57 & 81.46 & 73.36 & 63.73 & 84.73 & \textbf{72.75 (0.19)} \\\cline{2-15}
        &FixBi & 57.36 & 81.50 & 79.57 & 65.55 & 81.64 & 78.79 & 65.96 & 59.50 & 80.15 & 70.52 & 60.41 & 84.57 & \textbf{72.13 (0.16)} \\
        &SS-TrBoosting & 57.37 & 81.51 & 79.58 & 65.55 & 81.63 & 78.78 & 65.98 & 59.50 & 80.15 & 70.52 & 60.44 & 84.58 & \textbf{72.13 (0.16)}
\\\cline{2-15}
        &DNA & 55.95  & 78.02  & 78.34  & 63.74  & 76.23  & 74.09  & 62.71  & 55.14  & 78.98  & 70.92  & 59.25  & 83.38  & 69.73 (0.48) \\
        &SS-TrBoosting & 58.79  & 80.87  & 80.36  & 66.87  & 79.42  & 78.01  & 66.65  & 58.15  & 81.58  & 72.07  & 60.77  & 85.14  & \textbf{72.39 (0.44)}
\\\cline{2-15}
		&MCC   & 61.25 & 81.56 & 81.47 & 68.07 & 81.09 & 80.27 & 66.61 & 60.34 & 81.85 & 70.64 & 62.65 & 84.84 & 73.39 (0.14) \\
		&SS-TrBoosting & 61.54 & 81.64 & 81.53 & 68.12 & 81.22 & 80.24 & 66.67 & 60.36 & 82.19 & 70.61 & 62.70 & 84.81 & \textbf{73.47 (0.14)} \\\hline
		\multirow{6}{*}{SSDA}
		&ENT   & 56.68 & 79.19 & 79.48 & 67.74 & 79.69 & 77.35 & 66.37 & 56.67 & 80.17 & 72.80 & 59.71 & 83.55 & 71.62 (0.14) \\
		&SS-TrBoosting & 58.43 & 80.65 & 80.65 & 68.77 & 80.54 & 78.77 & 66.83 & 57.98 & 81.03 & 72.85 & 60.63 & 83.97 & \textbf{72.59 (0.09)} \\\cline{2-15}
		&MME   & 60.04 & 77.80 & 78.48 & 65.80 & 78.64 & 75.96 & 64.47 & 60.16 & 79.93 & 72.30 & 63.22 & 84.94 & 71.81 (0.25) \\
		&SS-TrBoosting & 60.50 & 78.75 & 79.70 & 66.29 & 79.15 & 76.85 & 64.65 & 60.29 & 80.62 & 72.39 & 63.21 & 85.15 & \textbf{72.30 (0.18)} \\\cline{2-15}
        &UODA  & 60.01 & 78.31 & 78.00 & 64.16 & 78.17 & 75.28 & 64.55 & 61.07 & 79.15 & 71.86 & 62.98 & 83.95 & 71.46 (0.11) \\
        &SS-TrBoosting & 61.82 & 80.57 & 80.35 & 64.84 & 80.01 & 76.87 & 65.49 & 62.05 & 80.75 & 72.24 & 64.04 & 85.09 & \textbf{72.85 (0.25)}
\\\cline{2-15}
		&APE   & 61.14 & 81.58 & 79.88 & 69.25 & 81.04 & 78.62 & 70.62 & 62.81 & 82.35 & 74.85 & 65.43 & 85.60 & 74.43 (0.07) \\
		&SS-TrBoosting & 62.21 & 82.10 & 80.91 & 69.79 & 81.33 & 79.39 & 70.74 & 63.31 & 82.69 & 74.82 & 65.50 & 85.97 & \textbf{74.90 (0.11)} \\
		\bottomrule
	\end{tabular}%
	\label{tab:office-home}%
\end{table*}%

\begin{table*}[ht]
	\caption{Accuracies ($\%$) on Office-31 for 1-shot and 3-shot settings for SSDA, when SS-TrBoosting was used to fine-tune.}
	\renewcommand \arraystretch{1.1}  \small  \centering \setlength{\tabcolsep}{1.5mm}
	\scalebox{1}{
		\begin{tabular}{c|c|cccccc|c}
			\toprule
			\multicolumn{2}{c|}{Algorithm} & A$\rightarrow$D & A$\rightarrow$W & D$\rightarrow$A & D$\rightarrow$W & W$\rightarrow$A & W$\rightarrow$D & Avg (std) \\
			\hline
			\multicolumn{9}{c}{\textbf{1-shot}} \\
			\hline
			&ST    & 87.37 (0.46) & 82.81 (1.76) & 68.11 (1.37) & 97.12 (0.60) & 68.31 (0.59) & 99.00 (0.36) & 83.79 (0.11) \\
			&SS-TrBoosting & 90.22 (1.02) & 86.34 (1.07) & 71.33 (1.11) & 97.69 (0.34) & 72.27 (0.68) & 99.14 (0.17) & \textbf{86.17 (0.13)} \\
			\hline
			\multirow{8}{*}{UDA}
			&DAN   & 81.73 (0.88) & 79.89 (2.41) & 67.70 (1.49) & 96.16 (0.96) & 67.49 (0.58) & 98.14 (0.27) & 81.85 (0.13) \\
			&SS-TrBoosting & 82.58 (1.77) & 82.81 (2.29) & 71.18 (1.25) & 97.12 (0.53) & 71.81 (0.10) & 98.36 (0.10) & \textbf{83.98 (0.40)} \\\cline{2-9}
			&DANN  & 86.87 (1.34) & 84.08 (0.87) & 66.02 (0.75) & 96.77 (0.99) & 66.48 (0.45) & 99.29 (0.20) & 83.25 (0.42) \\
			&SS-TrBoosting & 88.51 (0.96) & 88.48 (0.83) & 70.91 (0.66) & 97.77 (0.81) & 70.76 (1.22) & 99.64 (0.10) & \textbf{86.01 (0.33)} \\\cline{2-9}
			&CDAN  & 88.79 (0.66) & 90.97 (0.64) & 68.88 (1.35) & 97.99 (0.51) & 69.44 (0.95) & 99.86 (0.20) & 85.99 (0.09) \\
			&SS-TrBoosting & 88.79 (0.66) & 91.27 (0.38) & 71.93 (1.49) & 98.34 (0.83) & 71.39 (0.66) & 99.86 (0.20) & \textbf{86.93 (0.23)} \\\cline{2-9}
            &FixBi & 86.72 (0.17) & 90.23 (0.77) & 70.99 (2.03) & 98.52 (0.25) & 69.19 (1.42) & 98.07 (0.46) & 85.62 (0.26) \\
            &SS-TrBoosting & 86.94 (0.30) & 90.36 (0.89) & 71.08 (2.00) & 98.65 (0.06) & 69.26 (1.44) & 98.14 (0.36) & \textbf{85.74 (0.23)}
\\\cline{2-9}
            &DNA & 88.08 (0.90) & 88.17 (0.81) & 69.53 (0.56) & 97.95 (0.06) & 69.25 (1.34) & 99.72 (0.10) & 85.45 (0.63) \\
            &SS-TrBoosting & 89.36 (1.23) & 89.75 (0.96) & 73.11 (0.28) & 98.21 (0.43) & 71.68 (1.22) & 99.72 (0.10) & \textbf{86.97 (0.70)}
\\\cline{2-9}
			&MCC   & 90.72 (1.24) & 92.58 (1.98) & 74.35 (0.46) & 97.64 (0.28) & 73.79 (0.94) & 98.86 (0.10) & 87.99 (0.47) \\
			&SS-TrBoosting & 90.79 (1.15) & 92.58 (1.98) & 74.43 (0.46) & 97.69 (0.27) & 74.10 (1.14) & 98.86 (0.10) & \textbf{88.07 (0.46)} \\
			\hline
			\multirow{6}{*}{SSDA}
			&ENT   & 90.08 (0.83) & 90.18 (1.36) & 69.96 (0.33) & 97.56 (0.40) & 68.92 (1.15) & 99.29 (0.20) & 86.00 (0.12) \\
			&SS-TrBoosting & 90.36 (0.93) & 90.36 (1.50) & 72.05 (1.09) & 97.91 (0.39) & 71.64 (0.74) & 99.43 (0.27) & \textbf{86.96 (0.28)} \\\cline{2-9}
			&MME   & 88.51 (0.36) & 88.53 (1.73) & 65.72 (0.41) & 97.29 (1.12) & 68.56 (1.43) & 99.79 (0.17) & 84.73 (0.31) \\
			&SS-TrBoosting & 88.58 (0.40) & 88.53 (1.73) & 72.58 (0.65) & 98.08 (1.02) & 72.59 (0.93) & 99.93 (0.10) & \textbf{86.71 (0.24)} \\\cline{2-9}
            &UODA  & 85.87 (0.87) & 87.61 (0.79) & 65.75 (0.47) & 96.20 (0.64) & 66.01 (1.35) & 98.86 (0.53) & 83.38 (0.51) \\
            &SS-TrBoosting & 87.51 (1.59) & 89.97 (0.54) & 72.27 (0.07) & 97.91 (0.43) & 72.18 (0.75) & 99.64 (0.27) & \textbf{86.58 (0.47)}
\\\cline{2-9}
			&APE   & 92.51 (0.63) & 92.10 (1.74) & 70.76 (1.86) & 97.51 (1.11) & 72.25 (0.69) & 99.86 (0.10) & 87.50 (0.43) \\
			&SS-TrBoosting & 92.43 (0.53) & 92.10 (1.74) & 73.94 (1.64) & 97.64 (1.02) & 73.86 (0.70) & 99.93 (0.10) & \textbf{88.32 (0.44)} \\
			\hline
			\multicolumn{9}{c}{\textbf{3-shot}} \\
			\hline
			&ST    & 92.51 (0.81) & 90.88 (1.11) & 72.91 (0.77) & 97.34 (0.71) & 72.67 (0.62) & 99.42 (0.31) & 87.62 (0.25) \\
			&SS-TrBoosting & 94.49 (0.71) & 92.69 (1.71) & 75.93 (0.28) & 97.96 (0.82) & 75.73 (0.47) & 99.42 (0.31) & \textbf{89.37 (0.25)} \\
			\hline
			\multirow{8}{*}{UDA}
			&DAN   & 82.63 (1.11) & 84.71 (2.00) & 72.31 (0.92) & 96.91 (0.67) & 71.32 (0.84) & 98.02 (0.70) & 84.32 (0.37) \\
			&SS-TrBoosting & 86.17 (0.20) & 87.70 (2.55) & 75.18 (0.16) & 97.96 (0.37) & 75.09 (1.00) & 98.44 (0.12) & \textbf{86.76 (0.44)} \\\cline{2-9}
			&DANN  & 91.77 (1.83) & 91.12 (2.27) & 70.69 (0.93) & 97.86 (1.03) & 70.89 (1.49) & 99.34 (0.31) & 86.95 (0.37) \\
			&SS-TrBoosting & 93.99 (1.37) & 93.07 (2.15) & 74.47 (1.29) & 98.72 (1.11) & 75.55 (1.18) & 99.67 (0.12) & \textbf{89.25 (0.57)} \\\cline{2-9}
			&CDAN  & 94.65 (1.11) & 95.11 (1.08) & 73.58 (0.40) & 98.86 (0.53) & 74.49 (0.79) & 99.92 (0.12) & 89.43 (0.08) \\
			&SS-TrBoosting & 94.81 (1.52) & 95.16 (1.12) & 75.23 (0.36) & 98.86 (0.53) & 76.21 (0.50) & 100.00 (0.00) & \textbf{90.05 (0.06)} \\\cline{2-9}
            &FixBi & 91.60 (0.88) & 91.79 (1.47) & 76.68 (0.58) & 98.81 (0.86) & 76.41 (1.01) & 98.27 (0.53) & 88.93 (0.16) \\
            &SS-TrBoosting & 91.77 (0.99) & 91.93 (1.56) & 76.88 (0.52) & 98.81 (0.86) & 76.49 (1.03) & 98.44 (0.65) & \textbf{89.05 (0.22)}
\\\cline{2-9}
            &DNA & 90.70 (2.29) & 91.45 (1.18) & 74.24 (0.16) & 98.67 (0.55) & 74.01 (0.93) & 99.83 (0.12) & 88.15 (0.87) \\
            &SS-TrBoosting & 92.76 (2.07) & 92.78 (0.77) & 76.27 (0.20) & 98.58 (0.35) & 75.86 (1.30) & 99.67 (0.23) & \textbf{89.32 (0.82)}
\\\cline{2-9}
			&MCC   & 91.60 (1.76) & 93.07 (1.25) & 77.45 (0.36) & 98.20 (0.71) & 75.81 (0.99) & 97.78 (0.40) & 88.98 (0.36) \\
			&SS-TrBoosting & 91.60 (1.76) & 93.07 (1.25) & 77.73 (0.32) & 98.24 (0.77) & 76.02 (0.79) & 97.78 (0.40) & \textbf{89.07 (0.33)} \\
			\hline
			\multirow{6}{*}{SSDA}
			&ENT   & 93.25 (0.71) & 93.68 (0.98) & 74.51 (1.15) & 98.72 (0.35) & 73.41 (1.05) & 99.34 (0.23) & 88.82 (0.40) \\
			&SS-TrBoosting & 93.17 (0.71) & 93.78 (0.94) & 75.83 (0.76) & 98.77 (0.59) & 76.00 (0.52) & 99.42 (0.12) & \textbf{89.50 (0.23)} \\\cline{2-9}
			&MME   & 91.93 (1.11) & 94.16 (0.62) & 72.48 (0.93) & 97.72 (1.43) & 74.12 (0.45) & 99.75 (0.20) & 88.36 (0.10) \\
			&SS-TrBoosting & 91.93 (1.11) & 94.16 (0.62) & 75.83 (0.86) & 97.96 (1.19) & 76.32 (0.60) & 99.75 (0.20) & \textbf{89.33 (0.06)} \\\cline{2-9}
            &UODA  & 91.85 (0.73) & 91.60 (0.53) & 72.48 (0.89) & 97.72 (0.81) & 73.58 (0.54) & 98.93 (0.12) & 87.69 (0.44) \\
            &SS-TrBoosting & 92.18 (1.01) & 92.21 (0.18) & 76.65 (0.17) & 98.43 (0.60) & 76.31 (0.83) & 99.26 (0.00) & \textbf{89.17 (0.31)}
\\\cline{2-9}
			&APE   & 94.90 (0.76) & 94.78 (1.45) & 77.23 (0.70) & 98.20 (1.15) & 77.30 (0.76) & 99.92 (0.12) & 90.39 (0.29) \\
			&SS-TrBoosting & 94.90 (0.81) & 94.92 (1.35) & 78.23 (0.34) & 98.34 (1.25) & 77.99 (0.49) & 100.00 (0.00) & \textbf{90.73 (0.11)} \\
			\bottomrule
	\end{tabular}}%
	\label{tab:office-31}%
\end{table*}%

\subsubsection{DomainNet}

Table~\ref{tab:domainNet} shows that SS-TrBoosting can improve the performance of all DA approaches on DomainNet. Specifically, on average SS-TrBoosting improved the four SSDA baselines by 1.5$\%$  (0.8$\%$) in 1-shot (3-shot) learning. It also improved the UDA baselines by about 2.3$\%$ on average. Considering that DomainNet is a large dataset with class-imbalance and noisy labels, these results indicated that SS-TrBoosting is effective for handling large datasets, imbalanced classes, and noisy labels.

It is also interesting to note that some UDA approaches, e.g., CDAN and MCC, even outperformed SSDA approaches in 1-shot learning. This may be because only one labeled sample per class provided little information for this large dataset.

\subsubsection{Office-Home and Office-31}

As shown in Tables~\ref{tab:office-home} and \ref{tab:office-31}, SS-TrBoosting improved the performances of all baselines by over 1.5$\%$ on average on Office-Home and Office-31, respectively, demonstrating its effectiveness on medium-sized and small-sized datasets. Moreover, UDA's fine-tuned versions outperformed the SSDA baselines sometimes, and even achieved comparable performance with SSDA baselines' fine-tuned versions on Office-31, indicating that SS-TrBoosting can effectively extend UDA approaches to SSDA.


\subsection{SS-SFDA Results}

We transformed the SS-SFDA problem to an SSDA problem by using the pre-trained SFDA classifiers to generate a synthesized source domain.

\subsubsection{DomainNet}

As shown in Table~\ref{tab:SFDA_domainNet}, SS-TrBoosting combined with our proposed source generation approach improved all three baselines' performances on DomainNet for both 1-shot and 3-shot settings.

NRC did not perform well on DomainNet. This may be because NRC relies on the clustering assumption, which may not hold on datasets with high class-imbalance and noisy labels like DomainNet. SS-TrBoosting significantly improved the performance of NRC, though this result was not good, compared with the other two baselines. It suggests that SS-TrBoosting and the proposed source generation approach are robust to class-imbalance and noisy labels, but they also rely on the initial performance of baselines.

\begin{table*}[htbp]
	\caption{Accuracies ($\%$) on DomainNet for 1-shot and 3-shot settings for SS-SFDA, when SS-TrBoosting was used to fine-tune.}
	\renewcommand\arraystretch{1.1}  \small  \centering \setlength{\tabcolsep}{1.5mm}
	\begin{tabular}{c|c|cccccccccccc|c}
		\toprule
		\multicolumn{2}{c|}{Algorithm} & C$\rightarrow$P & C$\rightarrow$R & C$\rightarrow$S & P$\rightarrow$C & P$\rightarrow$R & P$\rightarrow$S & R$\rightarrow$C & R$\rightarrow$P & R$\rightarrow$S & S$\rightarrow$C & S$\rightarrow$P & S$\rightarrow$R & Avg (std) \\
		\hline
		\multicolumn{15}{c}{\textbf{1-shot}} \\
		\hline
		&SourceOnly & 42.27 & 59.14 & 44.72 & 46.60 & 61.34 & 41.29 & 52.16 & 49.88 & 41.11 & 55.69 & 46.14 & 58.03 & 49.86 \\
		&SS-TrBoosting & 45.45 & 62.55 & 45.74 & 50.13 & 64.51 & 42.36 & 53.96 & 51.51 & 42.53 & 57.47 & 48.00 & 61.27 & \textbf{52.12} \\
		\hline
		\multirow{4}{*}{SFDA}
		&SHOT  & 44.16 & 63.23 & 44.85 & 52.51 & 63.15 & 41.79 & 57.27 & 50.52 & 44.26 & 58.40 & 44.67 & 61.59 & 52.20 \\
		&SS-TrBoosting & 44.40 & 63.12 & 44.78 & 52.99 & 63.22 & 41.69 & 57.33 & 50.58 & 44.36 & 58.25 & 45.00 & 61.70 & \textbf{52.28} \\\cline{2-15}
		&NRC   & 23.34 & 27.42 & 26.38 & 30.79 & 32.39 & 21.47 & 28.47 & 24.28 & 20.84 & 31.92 & 22.75 & 28.48 & 26.54 \\
		&SS-TrBoosting & 31.25 & 35.41 & 29.38 & 34.89 & 40.54 & 27.03 & 38.67 & 35.31 & 28.01 & 33.45 & 29.16 & 29.94 & \textbf{32.75} \\
		\hline
		\multicolumn{15}{c}{\textbf{3-shot}} \\
		\hline
		&SourceOnly & 46.81 & 63.01 & 47.79 & 52.55 & 65.17 & 45.31 & 56.59 & 52.62 & 45.74 & 58.89 & 48.68 & 61.88 & 53.75 \\
		&SS-TrBoosting & 49.22 & 65.03 & 48.37 & 55.39 & 67.01 & 46.26 & 57.47 & 53.46 & 46.44 & 59.86 & 50.20 & 64.15 & \textbf{55.24} \\
		\hline
		\multirow{4}{*}{SFDA}
		&SHOT  & 44.54 & 63.41 & 45.17 & 52.83 & 63.27 & 42.26 & 57.42 & 50.66 & 44.37 & 58.57 & 45.18 & 61.91 & 52.47 \\
		&SS-TrBoosting & 45.49 & 64.05 & 45.84 & 53.82 & 63.80 & 42.85 & 57.79 & 51.09 & 44.70 & 59.19 & 46.23 & 62.35 & \textbf{53.10} \\\cline{2-15}
		&NRC   & 29.13 & 36.02 & 29.25 & 33.86 & 42.09 & 26.91 & 34.43 & 31.21 & 26.23 & 35.38 & 28.19 & 41.83 & 32.88 \\
		&SS-TrBoosting & 37.92 & 43.53 & 36.36 & 43.50 & 48.02 & 33.76 & 44.64 & 42.58 & 34.21 & 44.31 & 38.63 & 44.26 & \textbf{40.98} \\
		\bottomrule
	\end{tabular}%
	\label{tab:SFDA_domainNet}%
\end{table*}

\subsubsection{Office-Home and Office-31}

As shown in Tables~\ref{tab:SFDA_office-home} and \ref{tab:SFDA_office-31}, SS-TrBoosting improved SFDA baselines in almost all tasks on Office-Home and Office-31, in both 1-shot and 3-shot settings, demonstrating the effectiveness of SS-TrBoosting and our proposed source generation strategy.

In particular, SS-TrBoosting improved the average classification accuracy of SourceOnly by about 3.8$\%$ and 3.3$\%$ on Office-Home and Office-31, respectively, though SourceOnly provided neither domain invariant feature representations nor a transferable classifier. This suggests that SS-TrBoosting can transfer knowledge between two domains with large distribution discrepancies.

\begin{table*}[htbp]
	\caption{Accuracies ($\%$) on Office-Home for 1-shot and 3-shot settings for SS-SFDA, when SS-TrBoosting was used to fine-tune.}
	\renewcommand\arraystretch{1.1}  \small  \centering \setlength{\tabcolsep}{1.5mm}
	\begin{tabular}{c|c|cccccccccccc|c}
		\toprule
		\multicolumn{2}{c|}{Algorithm} & A$\rightarrow$C & A$\rightarrow$P & A$\rightarrow$R & C$\rightarrow$A & C$\rightarrow$P & C$\rightarrow$R & P$\rightarrow$A & P$\rightarrow$C & P$\rightarrow$R & R$\rightarrow$A & R$\rightarrow$C & R$\rightarrow$P & Avg (std) \\
		\hline
		\multicolumn{15}{c}{\textbf{1-shot}} \\
		\hline
		&SourceOnly & 46.98 & 67.26 & 74.07 & 57.82 & 66.64 & 68.33 & 57.25 & 45.74 & 73.88 & 67.25 & 49.99 & 77.76 & 62.75 (0.14) \\
		&SS-TrBoosting & 50.66 & 72.79 & 77.11 & 61.59 & 73.53 & 73.28 & 60.98 & 49.02 & 77.66 & 68.54 & 52.50 & 79.88 & \textbf{66.46 (0.08)} \\
		\hline
		\multirow{4}{*}{SFDA}
		&SHOT  & 55.70 & 78.65 & 81.30 & 68.76 & 77.72 & 78.46 & 68.15 & 53.64 & 82.54 & 72.61 & 57.67 & 83.75 & 71.58 (0.10) \\
		&SS-TrBoosting & 56.52 & 78.96 & 81.42 & 69.04 & 78.39 & 78.97 & 68.16 & 54.39 & 82.56 & 72.64 & 58.18 & 84.06 & \textbf{71.94 (0.08)} \\\cline{2-15}
		&NRC   & 57.69 & 78.84 & 80.02 & 66.55 & 78.28 & 78.44 & 67.25 & 57.08 & 81.07 & 70.63 & 59.11 & 82.02 & 71.42 (0.08) \\
		&SS-TrBoosting & 58.34 & 79.74 & 80.34 & 66.91 & 79.29 & 79.06 & 67.53 & 56.98 & 81.55 & 70.77 & 59.05 & 82.82 & \textbf{71.87 (0.06)} \\
		\hline
		\multicolumn{15}{c}{\textbf{3-shot}} \\
		\hline
		&SourceOnly & 49.42 & 71.45 & 74.90 & 60.93 & 71.76 & 70.68 & 60.36 & 50.19 & 75.37 & 69.35 & 53.22 & 79.32 & 65.58 (0.38) \\
		&SS-TrBoosting & 53.29 & 77.65 & 77.61 & 64.49 & 77.59 & 76.20 & 64.46 & 53.74 & 79.26 & 70.82 & 56.24 & 82.12 & \textbf{69.46 (0.21)} \\
		\hline
		\multirow{4}{*}{SFDA}
		&SHOT  & 56.45 & 78.75 & 81.55 & 69.31 & 78.51 & 78.89 & 68.83 & 54.43 & 82.72 & 73.24 & 58.24 & 83.84 & 72.06 (0.06) \\
		&SS-TrBoosting & 58.64 & 80.05 & 81.82 & 69.74 & 80.11 & 79.83 & 69.35 & 56.52 & 82.82 & 73.45 & 59.81 & 84.61 & \textbf{73.06 (0.09)} \\\cline{2-15}
		&NRC   & 60.70 & 80.51 & 79.14 & 67.08 & 80.44 & 79.18 & 68.18 & 60.75 & 81.13 & 71.40 & 61.87 & 83.46 & 72.82 (0.29) \\
		&SS-TrBoosting & 61.08 & 81.86 & 79.39 & 67.82 & 81.56 & 79.89 & 69.16 & 61.17 & 81.83 & 72.36 & 62.49 & 84.67 & \textbf{73.61 (0.23)} \\
		\bottomrule
	\end{tabular}%
	\label{tab:SFDA_office-home}%
\end{table*}%

\begin{table*}[htbp]
	\caption{Accuracies ($\%$) on Office-31 for 1-shot and 3-shot settings for SS-SFDA, when SS-TrBoosting was used to fine-tune.}
	\renewcommand \arraystretch{1.1}  \small  \centering \setlength{\tabcolsep}{1.5mm}
	\scalebox{1}{
		\begin{tabular}{c|c|cccccc|c}
			\toprule
			\multicolumn{2}{c|}{Algorithm} & A$\rightarrow$D & A$\rightarrow$W & D$\rightarrow$A & D$\rightarrow$W & W$\rightarrow$A & W$\rightarrow$D & Avg (std) \\
			\hline
			\multicolumn{9}{c}{\textbf{1-shot}} \\
			\hline
			&SourceOnly & 81.94 (0.50) & 79.41 (0.33) & 63.07 (0.63) & 96.47 (1.13) & 62.16 (0.25) & 99.14 (0.17) & 80.36 (0.14) \\
			&SS-TrBoosting & 85.30 (0.88) & 84.64 (1.73) & 70.02 (0.49) & 96.95 (0.99) & 68.21 (0.47) & 99.14 (0.17) & \textbf{84.04 (0.27)} \\
			\hline
			\multirow{4}{*}{SFDA}
			&SHOT  & 92.65 (1.70) & 89.92 (1.33) & 73.68 (1.12) & 97.08 (0.59) & 75.34 (0.65) & 99.29 (0.44) & 87.99 (0.55) \\
			&SS-TrBoosting & 93.00 (2.03) & 89.83 (1.28) & 74.07 (1.24) & 97.25 (0.83) & 75.44 (0.67) & 99.29 (0.44) & \textbf{88.15 (0.60)} \\\cline{2-9}
			&NRC   & 92.51 (1.06) & 91.49 (0.67) & 75.50 (1.58) & 97.43 (1.01) & 75.45 (1.08) & 98.14 (0.27) & 88.42 (0.27) \\
			&SS-TrBoosting & 92.86 (0.71) & 91.54 (0.63) & 75.47 (1.58) & 97.43 (1.01) & 75.42 (1.09) & 98.00 (0.20) & \textbf{88.45 (0.18)} \\
			\hline
			\multicolumn{9}{c}{\textbf{3-shot}} \\
			\hline
			&SourceOnly & 88.31 (1.03) & 88.27 (0.64) & 67.45 (0.97) & 97.91 (0.18) & 66.47 (1.24) & 99.59 (0.12) & 84.67 (0.36) \\
			&SS-TrBoosting & 90.78 (0.99) & 92.12 (0.24) & 73.13 (0.39) & 97.67 (0.37) & 71.79 (1.13) & 99.26 (0.60) & \textbf{87.46 (0.50)} \\
			\hline
			\multirow{4}{*}{SFDA}
			&SHOT  & 93.33 (0.92) & 89.55 (1.34) & 73.86 (0.82) & 96.96 (0.47) & 75.40 (0.85) & 99.51 (0.20) & 88.10 (0.41) \\
			&SS-TrBoosting & 93.83 (0.73) & 90.22 (1.28) & 74.85 (0.87) & 97.25 (0.59) & 76.31 (0.73) & 99.51 (0.20) & \textbf{88.66 (0.40)} \\\cline{2-9}
			&NRC   & 93.66 (1.23) & 92.55 (0.82) & 78.17 (0.27) & 98.15 (0.42) & 77.25 (0.82) & 97.53 (0.60) & 89.55 (0.19) \\
			&SS-TrBoosting & 93.83 (1.12) & 92.64 (0.83) & 78.19 (0.35) & 98.20 (0.36) & 77.23 (0.84) & 97.53 (0.60) & \textbf{89.60 (0.21)} \\
			\bottomrule
	\end{tabular}}%
	\label{tab:SFDA_office-31}%
\end{table*}%


\subsection{Additional Analyses}

Next, we performed experiments to analyze the effects of random nonlinear feature mapping and misclassified source data removal, the sensitivity of SS-TrBoosting to the noise magnitude $\xi$ and the number of fine-tuning blocks $K$, and the quality of the synthesized source data for SS-SFDA. We also compare SS-TrBoosting with traditional semi-supervised boosting approach.

\begin{figure}[htbp]	\centering
	\subfigure[]{\label{fig:NewNode}   \includegraphics[width=0.4\textwidth,clip]{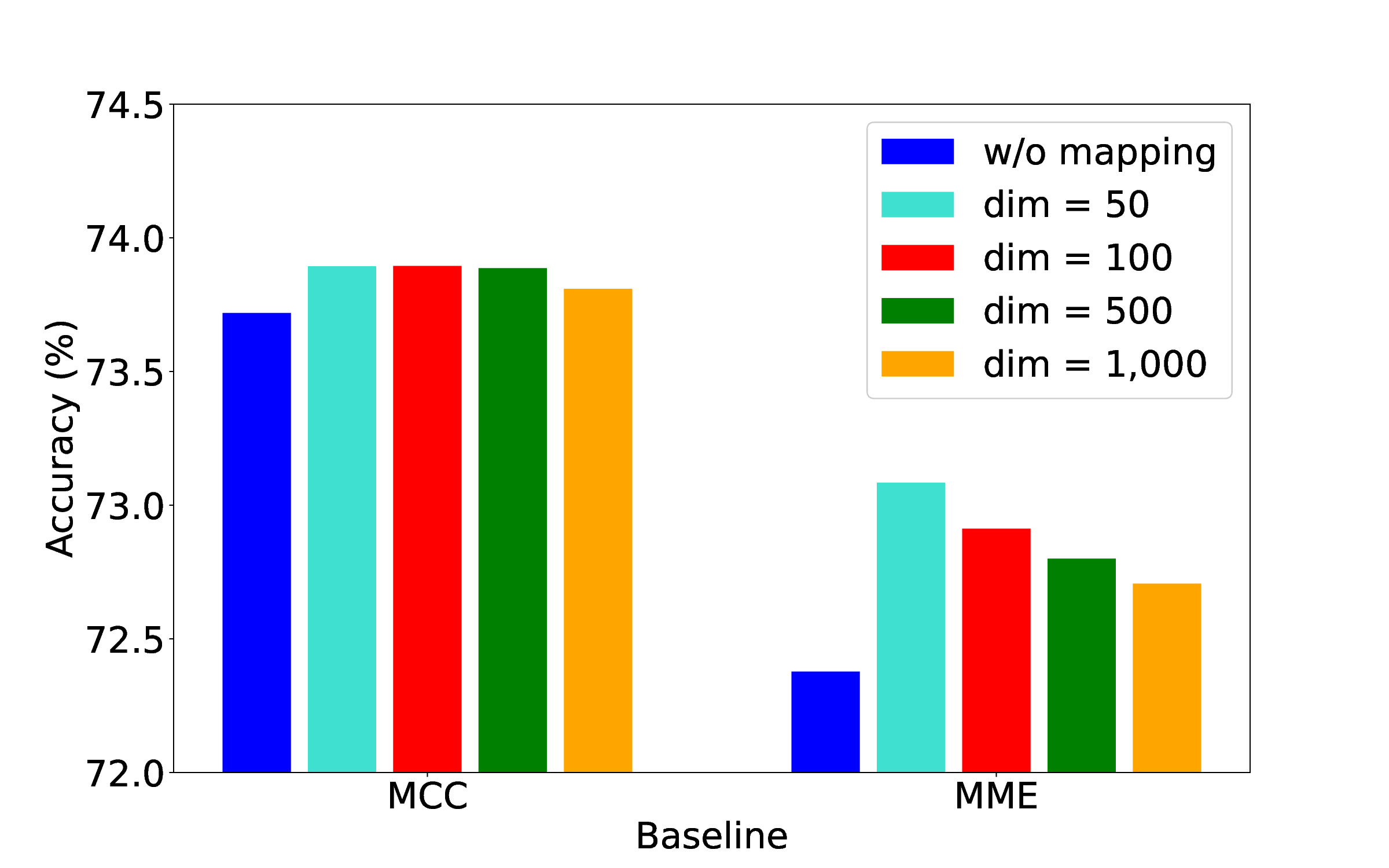}}
	\subfigure[]{\label{fig:Weight}    \includegraphics[width=0.4\textwidth,clip]{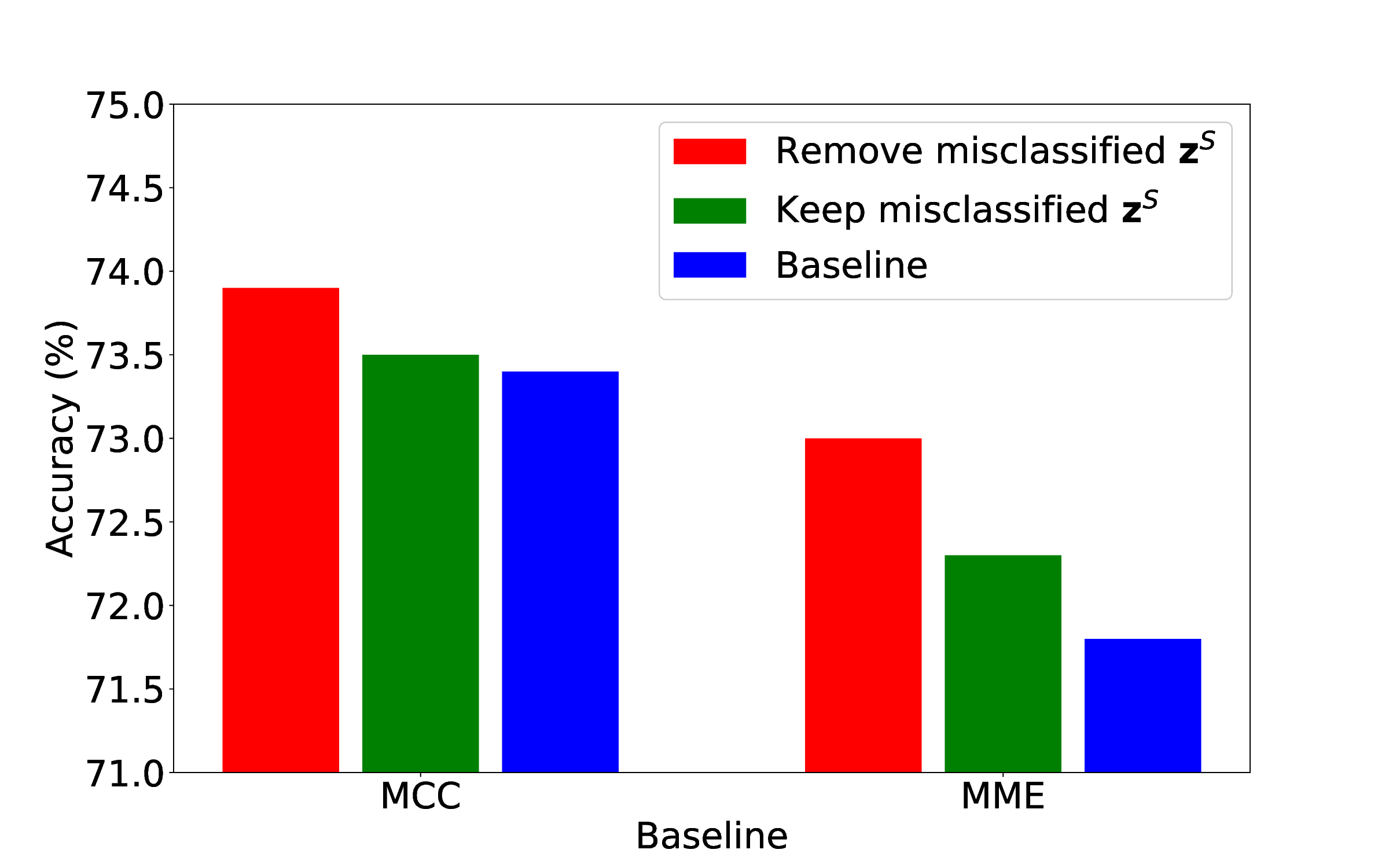}}
	\subfigure[]{\label{fig:Augment}   \includegraphics[width=0.4\textwidth,clip]{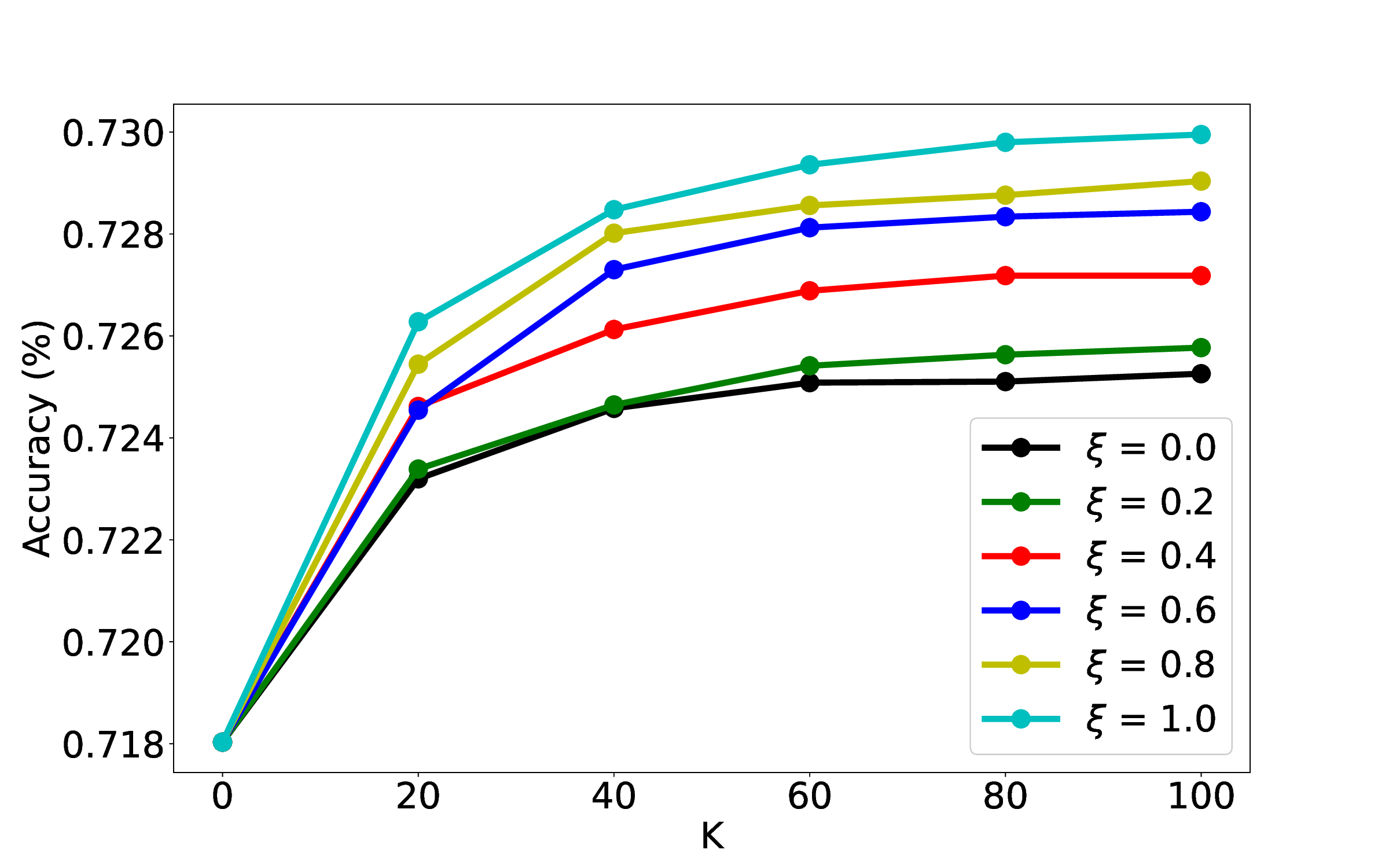}}
	\subfigure[]{\label{fig:Iteration} \includegraphics[width=0.4\textwidth,clip]{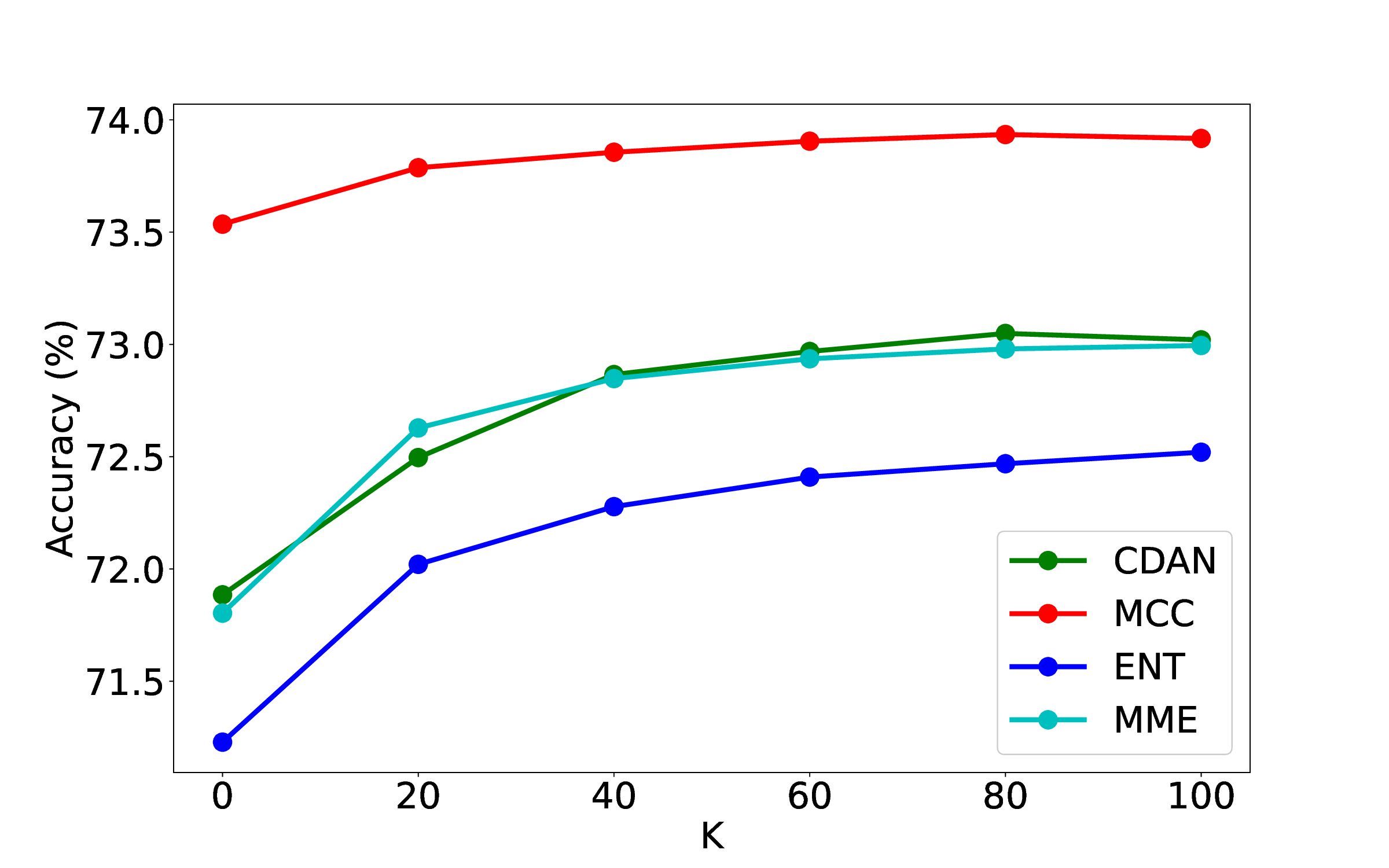}}
	\caption{(a) Effect of random nonlinear feature mapping. ``dim" is the dimensionality of the mapped features. (b) Effect of removing misclassified source data. $\bm{z}^S$ denotes source data. (c) Sensitivity to the noise magnitude $\xi$. The baseline was MME. (d) Sensitivity to the number of fine-tuning blocks $K$.}\label{fig:analysis}
\end{figure}

\subsubsection{Effect of random nonlinear feature mapping}

Fig.~\ref{fig:NewNode} shows the accuracies of MCC's and MME's fine-tuned versions with and without random nonlinear feature mapping, as the dimensionality of the mapped features increased from 50 to 1,000. Random nonlinear feature mapping can improve the performance of both baselines. However, mapping features to a high-dimensional space may hurt the performance.

\subsubsection{Effect of removing the misclassified source data}

Fig.~\ref{fig:Weight} shows the accuracies of MCC's and MME's fine-tuned versions with and without setting the weights of the misclassified source data to zero in supervised DA in each fine-tuning block. Removing the misclassified source data improved the performance of both baselines, implying that it helped reduce the domain alignment bias.

Without removing the misclassified source data, SS-TrBoosting still outperformed the baselines, suggesting its robustness.

\subsubsection{Sensitivity to the noise magnitude $\xi$}

Fig.~\ref{fig:Augment} shows the accuracies of SS-TrBoosting in fine-tuning MME (baseline), as $\xi$ increased from 0 to 1.0 ($\xi=0$ means no data augmentation was used in SS-TrBoosting). Adding noise to the unlabeled target data can improve the performance of SS-TrBoosting and accelerate its convergence. The bigger the noise (i.e., higher value of $\xi$, $\xi \leq 1.0$), the greater the improvement.

\subsubsection{Sensitivity to the number of fine-tuning blocks $K$}

Fig.~\ref{fig:Iteration} shows the accuracies of SS-TrBoosting in fine-tuning four baselines, as $K$ increased from 0 to 100. Generally, when $K$ exceeded 100, SS-TrBoosting converged for most UDA and SSDA baselines, but ENT had under-fitting. A possible reason is that ENT does not align the distributions of the source and target domains, leading to slow convergence.

\subsubsection{Comparison with traditional semi-supervised boosting}

We compared SS-TrBoosting with another boosting approach, ASSEMBLE \cite{bennett2002exploiting}, on five baselines. Both approaches took a baseline's extracted features and last linear layer as inputs and the initial base learner, respectively.

As shown in Table~\ref{tab:comparison_boost}, ASSEMBLE reduced the performance of all five baselines by over 4$\%$ on average, indicating that simply combining boosting with deep-learning-based DA approaches may not be effective.

\begin{table}[htbp]
 \renewcommand \arraystretch{1.1}  \small  \centering \setlength{\tabcolsep}{1.5mm}
  \caption{Average accuracies ($\%$) on Office-Home for 3-shot setting for SSDA, when ASSEMBLE and SS-TrBoosting were used to fine-tune different baselines. ``Original" means the original baseline results.}
    \begin{tabular}{c|ccccc}
    \toprule
    Algorithm & DAN   & CDAN  & MCC   & ENT   & MME \\
    \hline
    Original & 65.36 & 72.25 & 73.39 & 71.62 & 71.81 \\
    \hline
    ASSEMBLE & 61.14 & 65.60 & 70.32 & 67.40 & 67.70 \\
    \hline
    SS-TrBoosting & \textbf{68.87} & \textbf{72.75} & \textbf{73.47} & \textbf{72.59} & \textbf{72.31} \\
    \bottomrule
    \end{tabular}%
  \label{tab:comparison_boost}%
\end{table}%

\subsubsection{Computational cost of SS-TrBoosting}

We compared the computational cost of SS-TrBoosting and Simple Ensemble. SS-TrBoosting trained multiple linear models to fine-tune CDAN. Simple Ensemble trained multiple CDAN models with different initializations and took their average as the output.

The results are shown in Table~\ref{tab:time_cost}. The computational cost of Simple Ensemble was significantly larger than SS-TrBoosting, demonstrating the efficiency of our proposed approach.

\begin{table}[htpb]
 \renewcommand \arraystretch{1}  \small  \centering \setlength{\tabcolsep}{1.5mm}  
  \caption{Computational cost (seconds) of SS-TrBoosting and Simple Ensemble on three datasets for 3-shot SSDA. CDAN was the baseline model.}
    \begin{tabular}{c|ccc}
    \toprule
    Dataset                 & Office              & Office-Home          & DomainNet \\
    \hline
    SS-TrBoosting  & $\mathbf{9.63\times 10^2}$   & $\mathbf{2.47\times 10^3}$    & $\mathbf{3.53\times 10^4}$  \\
    \hline
    Simple Ensemble         & $6.92\times 10^4$   & $1.77\times 10^5$    & $1.67\times 10^6$  \\
    \bottomrule
    \end{tabular}%
  \label{tab:time_cost}%
\end{table}%

\subsubsection{Visualization of the synthesized source data}

We used t-SNE \cite{van2008visualizing} to visualize the representations of the synthesized source data and the true source data, which were generated by SHOT in task D$\rightarrow$W on Office-31 under 3-shot setting for SS-SFDA.

\begin{figure} [htpb]     \centering
    \includegraphics[width=.47\textwidth,clip]{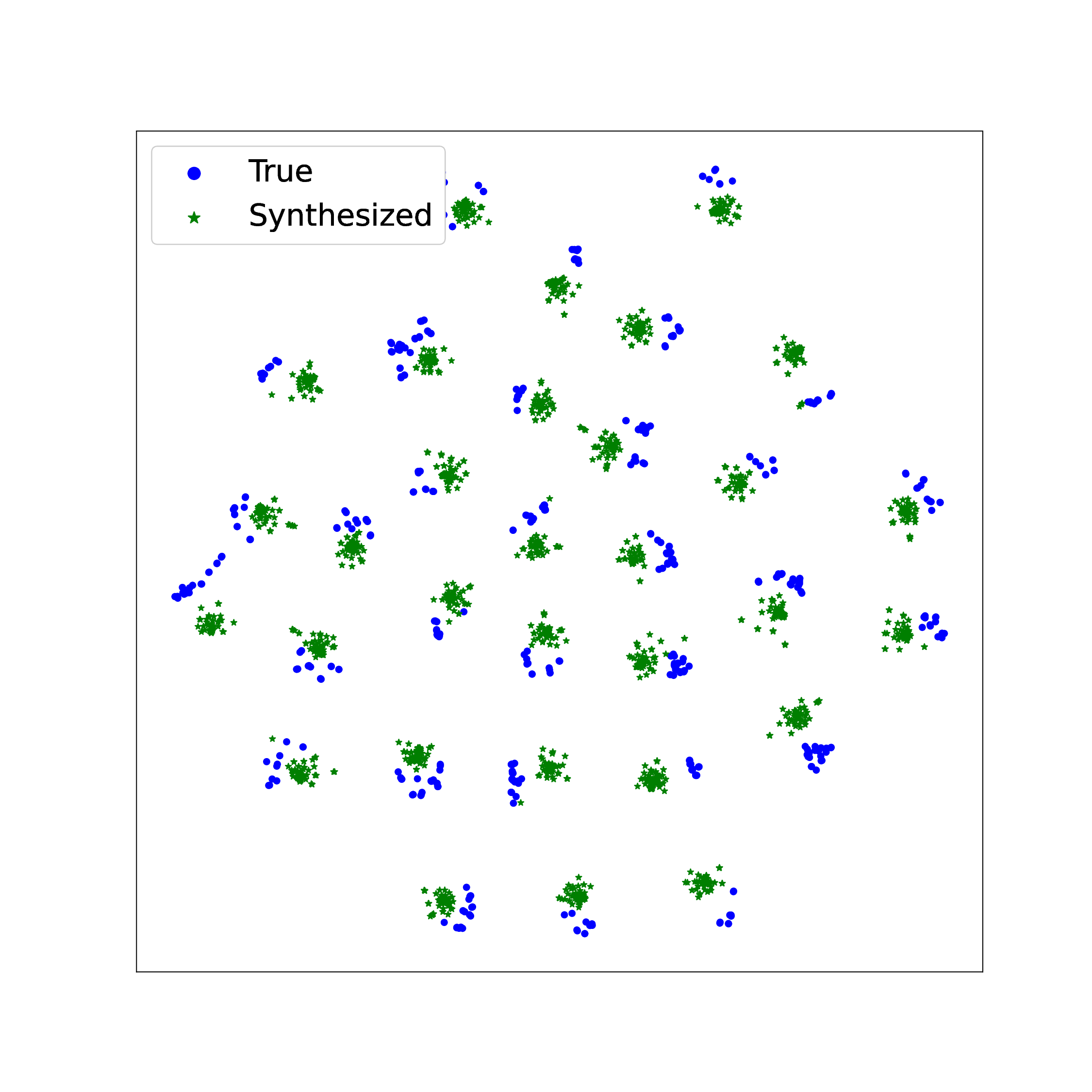}
    \caption{t-SNE feature visualization of the synthesized source data and the true source data for a 31-way classification task D$\rightarrow$W on Office-31.}
    \label{fig:source_generation}
\end{figure}

As shown in Fig.~\ref{fig:source_generation}, the two domains were well aligned, demonstrating that the synthesized source domain resembles the true source domain.

\section{Conclusions and future work}

This paper has proposed a novel transfer-boosting-based fine-tuning approach, SS-TrBoosting, for SSDA. SS-TrBoosting is compatible with a variety of deep-learning-based unsupervised or semi-supervised DA approaches, and can further improve their generalization performance. It first uses the DA model's feature extractor to pre-align the source and target domains, and then generates a series of fine-tuning blocks to enhance the DA model's classification performance. In each boosting fine-tuning block, SS-TrBoosting first decomposes the SSDA problem into a supervised DA problem and an SSL problem, and then trains a base learner for each of them. Furthermore, we also propose a novel source data synthesis approach to extend SS-TrBoosting to SS-SFDA. Extensive experiments demonstrated the effectiveness, flexibility and robustness of our proposed SS-TrBoosting.

In the future, we plan to apply SS-TrBoosting to more domain adaptation scenarios, such as unsupervised domain adaptation and few-shot domain adaptation \cite{motiian2017few}.


\end{document}